\documentclass[10pt]{article}
\usepackage[letterpaper,margin=1in]{geometry}
\usepackage[T1]{fontenc}
\usepackage[utf8]{inputenc}
\usepackage{newtxtext}
\usepackage{amsmath,amssymb,amsthm,mathtools,mathrsfs}
\usepackage{newtxmath}
\DeclareFontFamily{U}{rsfs}{\skewchar\font127}
\DeclareFontShape{U}{rsfs}{m}{n}{<-6>rsfs5 <6-8>rsfs7 <8->rsfs10}{}
\usepackage[scaled=0.92]{helvet}
\usepackage[varqu]{inconsolata}
\usepackage{aliascnt}
\usepackage{bm}
\usepackage{booktabs,longtable,array,multirow,tabularx}
\usepackage{graphicx}
\usepackage{tikz}
\usetikzlibrary{arrows.meta,positioning,fit,calc,backgrounds,shapes.geometric,decorations.pathreplacing,matrix}
\usepackage{xcolor}
\usepackage{microtype}
\usepackage{enumitem}
\usepackage{caption}
\usepackage{float}
\usepackage[numbers,sort&compress]{natbib}
\usepackage{titlesec}
\usepackage{xurl}
\usepackage{placeins}
\usepackage{setspace}
\usepackage[hidelinks]{hyperref}
\usepackage{doi}
\usepackage[nameinlink,capitalise,noabbrev]{cleveref}

\definecolor{FFBlue}{HTML}{4C78A8}
\definecolor{FFBlueLight}{HTML}{EAF1F8}
\definecolor{FFGold}{HTML}{D3A52F}
\definecolor{FFGoldLight}{HTML}{FBF4DF}
\definecolor{FFRed}{HTML}{B56470}
\definecolor{FFRedLight}{HTML}{F8ECEE}
\definecolor{FFGray}{HTML}{6A7179}
\definecolor{FFGrayDark}{HTML}{3F454B}
\definecolor{FFRule}{HTML}{D3D7DB}
\definecolor{FFLight}{HTML}{F7F8F9}
\definecolor{FFInk}{HTML}{111315}
\definecolor{RCTeal}{HTML}{2C8C87}
\definecolor{RCTealLight}{HTML}{E7F3F1}

\color{FFInk}
\setlist{leftmargin=*,topsep=3pt,itemsep=1.8pt,parsep=0pt,partopsep=0pt}
\newcolumntype{L}[1]{>{\raggedright\arraybackslash}p{#1}}
\newcolumntype{Y}{>{\raggedright\arraybackslash}X}
\hypersetup{
  pdfauthor={Qinyou Wang},
  pdftitle={Revelation Control},
  pdfsubject={A decision theory of priced interventions that reveal hidden decision-relevant state, with productive reuse and adaptive depth},
  pdfkeywords={revelation control, decision factorization, productive experiments, adaptive control, value of information, dual control, sequential decision making, hidden learning state},
  pdfcreator={pdfLaTeX with TikZ/PGF},
  pdfdisplaydoctitle=true
}
\titleformat{\section}{\large\bfseries}{\thesection}{0.68em}{}
\titleformat{\subsection}{\normalsize\bfseries}{\thesubsection}{0.60em}{}
\titleformat{\subsubsection}{\normalsize\bfseries\itshape}{\thesubsubsection}{0.52em}{}
\titlespacing*{\section}{0pt}{12pt plus 3pt minus 2pt}{4.5pt plus 1pt minus 1pt}
\titlespacing*{\subsection}{0pt}{8.5pt plus 2pt minus 1pt}{2.8pt plus 1pt minus 1pt}
\titlespacing*{\subsubsection}{0pt}{6.5pt plus 1.5pt minus 1pt}{2.2pt}

\newtheoremstyle{ffplain}{6pt}{6pt}{\itshape}{}{\bfseries}{.}{0.5em}{}
\newtheoremstyle{ffdefinition}{6pt}{6pt}{\normalfont}{}{\bfseries}{.}{0.5em}{}
\newtheoremstyle{ffremark}{5.5pt}{5.5pt}{\normalfont}{}{\itshape}{.}{0.5em}{}
\theoremstyle{ffplain}
\newtheorem{theorem}{Theorem}[section]
\newaliascnt{proposition}{theorem}\newtheorem{proposition}[proposition]{Proposition}\aliascntresetthe{proposition}
\newaliascnt{corollary}{theorem}\newtheorem{corollary}[corollary]{Corollary}\aliascntresetthe{corollary}
\newaliascnt{lemma}{theorem}\aliascntresetthe{lemma}
\theoremstyle{ffdefinition}
\newaliascnt{definition}{theorem}\newtheorem{definition}[definition]{Definition}\aliascntresetthe{definition}
\newaliascnt{assumption}{theorem}\aliascntresetthe{assumption}
\newaliascnt{protocol}{theorem}\newtheorem{protocol}[protocol]{Protocol}\aliascntresetthe{protocol}
\theoremstyle{ffremark}
\newaliascnt{remark}{theorem}\aliascntresetthe{remark}

\AddToHook{env/theorem/begin}{\crefalias{section}{theorem}}
\AddToHook{env/proposition/begin}{\crefalias{section}{proposition}}
\AddToHook{env/corollary/begin}{\crefalias{section}{corollary}}
\AddToHook{env/lemma/begin}{\crefalias{section}{lemma}}
\AddToHook{env/definition/begin}{\crefalias{section}{definition}}
\AddToHook{env/assumption/begin}{\crefalias{section}{assumption}}
\AddToHook{env/protocol/begin}{\crefalias{section}{protocol}}
\AddToHook{env/remark/begin}{\crefalias{section}{remark}}

\makeatletter
\renewcommand{\maketitle}{%
  \begin{center}
    {\fontsize{19}{22.5}\selectfont\bfseries \@title\par}
    \vspace{1.15em}
    {\normalsize \@author\par}
  \end{center}
  \vspace{0.50em}
}
\makeatother

\renewenvironment{abstract}{%
  \begin{center}\begin{minipage}{0.92\linewidth}\small
  \begin{center}\bfseries Abstract\end{center}\vspace{-0.45em}
}{%
  \end{minipage}\end{center}\vspace{0.35em}
}

\tikzset{
  ffbox/.style={
    draw=FFGrayDark,
    line width=0.64pt,
    rounded corners=1.25pt,
    align=center,
    inner xsep=2.7mm,
    inner ysep=1.7mm,
    font=\sffamily\footnotesize,
    text=FFInk
  },
  ffneutral/.style={ffbox,fill=FFLight},
  ffblue/.style={ffbox,fill=FFBlueLight,draw=FFBlue!80!black},
  ffgold/.style={ffbox,fill=FFGoldLight,draw=FFGold!75!black},
  ffred/.style={ffbox,fill=FFRedLight,draw=FFRed!78!black},
  ffteal/.style={ffbox,fill=RCTealLight,draw=RCTeal!80!black},
  ffwhite/.style={ffbox,fill=white},
  ffgray/.style={ffbox,fill=FFLight,draw=FFRule},
  ffdecision/.style={
    diamond,
    aspect=2.25,
    draw=FFGrayDark,
    fill=white,
    line width=0.66pt,
    inner xsep=1.7mm,
    inner ysep=1.2mm,
    align=center,
    font=\sffamily\scriptsize,
    text=FFInk
  },
  ffpanelbox/.style={
    draw=FFRule,
    fill=white,
    rounded corners=1.5pt,
    line width=0.52pt
  },
  ffdashframe/.style={
    draw=FFGray,
    line width=0.58pt,
    rounded corners=1.3pt,
    densely dashed,
    fill=FFLight!55,
    inner sep=2.4mm,
    align=center,
    font=\sffamily\footnotesize,
    text=FFInk
  },
  ffarrow/.style={
    -{Latex[length=1.9mm,width=1.12mm]},
    draw=FFGrayDark,
    line width=0.70pt,
    rounded corners=1.0pt
  },
  ffarrowblue/.style={
    -{Latex[length=1.9mm,width=1.12mm]},
    draw=FFBlue,
    line width=0.80pt,
    rounded corners=1.0pt
  },
  ffarrowgold/.style={
    -{Latex[length=1.9mm,width=1.12mm]},
    draw=FFGold!90!black,
    line width=0.80pt,
    rounded corners=1.0pt
  },
  ffarrowred/.style={
    -{Latex[length=1.9mm,width=1.12mm]},
    draw=FFRed,
    line width=0.80pt,
    rounded corners=1.0pt
  },
  ffarrowteal/.style={
    -{Latex[length=1.9mm,width=1.12mm]},
    draw=RCTeal,
    line width=0.80pt,
    rounded corners=1.0pt
  },
  ffdasharrow/.style={
    -{Latex[length=1.72mm,width=1.04mm]},
    draw=FFGray,
    line width=0.58pt,
    densely dashed,
    rounded corners=1.0pt
  },
  ffline/.style={draw=FFGrayDark,line width=0.64pt},
  ffrule/.style={draw=FFRule,line width=0.54pt},
  ffdashline/.style={draw=FFGray,line width=0.56pt,densely dashed},
  ffpanel/.style={font=\sffamily\scriptsize\bfseries,text=FFGrayDark,anchor=west},
  ffpanelaccent/.style={font=\sffamily\scriptsize\bfseries,text=FFInk,anchor=west},
  fflabel/.style={font=\sffamily\scriptsize,text=FFInk,align=center},
  ffsubtle/.style={font=\sffamily\tiny,text=FFGray,align=center},
  ffmath/.style={font=\footnotesize,text=FFInk,align=center},
  ffdot/.style={circle,minimum size=2.5pt,inner sep=0pt,draw=none,fill=FFGrayDark}
}

\crefname{theorem}{Theorem}{Theorems}\Crefname{theorem}{Theorem}{Theorems}
\crefname{proposition}{Proposition}{Propositions}\Crefname{proposition}{Proposition}{Propositions}
\crefname{corollary}{Corollary}{Corollaries}\Crefname{corollary}{Corollary}{Corollaries}
\crefname{lemma}{Lemma}{Lemmas}\Crefname{lemma}{Lemma}{Lemmas}
\crefname{definition}{Definition}{Definitions}\Crefname{definition}{Definition}{Definitions}
\crefname{assumption}{Assumption}{Assumptions}\Crefname{assumption}{Assumption}{Assumptions}
\crefname{protocol}{Protocol}{Protocols}\Crefname{protocol}{Protocol}{Protocols}
\crefname{remark}{Remark}{Remarks}\Crefname{remark}{Remark}{Remarks}

\numberwithin{equation}{section}
\newcommand{\HB}{\mathscr H}
\newcommand{\GB}{\mathscr G}

\newcommand{\E}{\mathbb E}
\newcommand{\Pp}{\mathbb P}
\newcommand{\ind}{\mathbf 1}
\newcommand{\KEEP}{\textnormal{\textsc{Keep}}}
\newcommand{\XI}{\textnormal{\textsc{Xi}}}
\newcommand{\AR}{\textnormal{\textsc{AR}}}
\newcommand{\Frontier}{\textnormal{\textsc{Frontier}}}
\newcommand{\Vpol}{\mathcal V}

\title{Revelation Control}
\author{Qinyou Wang}
\date{}

\begin{document}
\maketitle
\thispagestyle{empty}
\begin{abstract}
Revelation Control is the problem of choosing priced interventions that reveal hidden state only insofar as the revealed distinctions can change a consequential decision, while accounting separately for any useful progress created by the intervention itself. We develop this theory for learning systems, where states equivalent under declared current information can respond differently to future training and favor different actions. The framework defines decision-sufficient revelation and revelation depth, separates pure information value from productive reuse, embeds static Bayes refinement into state-dependent continuation value, and gives an exact cost-adjusted factorization criterion: an additional shallow coordinate is decision-nonredundant only when states sharing a scalar summary lie on opposite sides of the priced \textsc{Stop}/\textsc{Continue} boundary. We also give a target-independent protocol for model-specific instantiation and prove that bounded stop--flip risk alone cannot certify positive expected utility under unrestricted severity. Across Qwen2.5-7B and Mistral-7B-v0.3, deeper future-learning probes have positive decision value and productive reuse yields strict equal-compute utility advantages. Qwen additionally provides evidence for a decision-nonredundant shallow revealability regime; in Mistral, a scalar continuation architecture fit only on an independent development panel retains positive familywise-adjusted lower bounds on a disjoint target panel, consistent with scalar decision sufficiency within the tested architecture family and resolution. The evidence supports structural rather than numerical transfer: the decision theory, cost accounting, continuation logic, and evaluation protocol transport, while empirical proxies, coefficients, thresholds, and even the required shallow state dimension may be system-specific.
\end{abstract}

\section{Introduction}

Sequential decision systems are routinely controlled through summaries of their current state. In learning systems these include present losses, held-out metrics, gradients, optimizer telemetry, and short diagnostic trajectories. Such summaries are decision-sufficient only when the distinctions they discard cannot change the action that should be taken. A learner may instead occupy execution states that look equivalent under the declared current information yet react differently to the same future training intervention.

Our companion work, \emph{Fiber Fingerprints of Hidden Learning-State Dynamics} \citep{wang2026fiber}, establishes the predictive premise within its declared scope: present behavior need not be a sufficient statistic for declared future learning. The decision theory developed here does not depend on that paper's specific geometric machinery; it takes hidden decision-relevant state as the object to be revealed and asks the downstream question:
\begin{quote}
\emph{When does hidden learning-state structure become actionable decision information?}
\end{quote}
We call the resulting control problem \emph{Revelation Control}: the controller chooses not only which action to take, but which future-learning interventions to instantiate, how much decision-relevant state to reveal, and which tested future to promote into execution.

The distinction between prediction and decision is essential. A hidden variable can improve prediction without changing the optimal action; a refined observation can increase oracle Bayes value while a finite-sample learner fails to extract it; and a training probe can both reveal information and leave behind useful computation. A decision theory for future-learning probes must therefore keep information, execution technology, learnability, and cost separate. Once refinement depth is allowed to vary, the same value functional becomes a control law over how much future information to acquire.

Partial training itself is not new. Freeze--Thaw Bayesian optimization uses partial learning curves to decide whether to pause or resume models \citep{swersky2014freeze}; learning-curve extrapolation can terminate weak runs before full training \citep{domhan2015speeding}; successive resource-allocation methods exploit intermediate performance to concentrate computation on promising candidates \citep{jamieson2016nonstochastic,li2018hyperband}; Population Based Training reuses and adapts live training trajectories \citep{jaderberg2017pbt}; and DynaMiCS uses short domain-specific probes to estimate local cross-domain effects before selecting a constrained fine-tuning mixture \citep{gualdoni2026dynamics}. Nor are Bayes value of information, Bayesian experimental design, active sensing, partial observability, or dual control new ideas \citep{blackwell1953equivalent,lindley1956measure,chaloner1995bayesian,howard1966information,feldbaum1960dual,kaelbling1998planning,veiga2023active}. The distinctive step here is to turn \emph{future learning itself} into a decision-relative experiment on hidden learner state, and to exploit the fact that the tested path can remain useful computation. Although the empirical instantiation in this paper is model training, the underlying decision structure is broader: a priced, controlled probe can change a system, reveal otherwise hidden decision-relevant state through its response, and leave reusable progress if the tested path is selected. \Cref{sec:applications} develops this extension while keeping all empirical claims confined to the learning systems actually studied.

\subsection{Contributions}
\label{sec:contributions}
The paper makes five coupled contributions.
\begin{enumerate}[label=\textbf{\arabic*.}]
\item \textbf{Decision-sufficient revelation.} We characterize when refined information changes the Bayes decision quotient, derive exact binary aliasing value, and identify a local revelation depth at which currently hidden decision-relevant directions become observable to an admissible future-learning probe.
\item \textbf{Productive revelation technology.} We distinguish discard-and-restart probing from probe-and-promote revelation, derive the equal-budget depth geometry, and separate pure information gain from the value of having already executed part of the selected training path.
\item \textbf{Adaptive revelation control.} We prove that population Bayes refinement value is exactly the expectation of a conditional state-dependent revelation value. We then derive an exact cost-adjusted scalar-control gap: a second shallow coordinate is decision-nonredundant precisely when a positive-mass scalar fiber contains states on both sides of the priced \textsc{Stop}/\textsc{Continue} boundary. In the binary location--scale specialization this yields a critical-revealability crossing criterion, so one-dimensional control can remain optimal even when revealability varies conditionally but never changes the meta-action.
\item \textbf{Learnability and certification boundaries.} We separate oracle revelation value from approximation, estimation, acquisition, and promotion terms. For adaptive depth, we identify a sharp boundary: bounded stop--flip risk alone cannot certify positive expected utility under unrestricted severity; a severity, moment, or integrable-tail condition is the missing object.
\item \textbf{Cross-model Transformer validation and a reusable instantiation procedure.} In a first 7B Transformer family, fixed-depth revelation, structural decision refinement, productive reuse, equal-compute frontier advantage, and adaptive safe-compute behavior are supported on independent panels. In an independently instantiated Mistral-7B-v0.3 family, positive deeper-revelation value and productive equal-compute utility reproduce on an independent two-bank panel. Within the finite pre-existing continuation-architecture family, a scalar architecture fit only on separate development data retains positive familywise-adjusted lower bounds on the target panel. The two families therefore reproduce the same Revelation-Control structure, while Qwen provides evidence that extra shallow revealability information is decision-nonredundant at the declared compute price and Mistral is consistent with scalar decision sufficiency within the tested architecture family and resolution.
\end{enumerate}

\subsection{Scope and organization}
\label{sec:scope-organization}
\paragraph{Scope of claims.}
The paper does not claim unrestricted dominance over every current-information policy, every probing algorithm, every model family, or every deployment cost model. The finite-library result is exactly that: finite-library. The structural refinement claim is conditional on a declared shallow ambiguity regime rather than an unrestricted Bayes comparison over complete shallow information. The comparative claim uses one strengthened DynaMiCS-style short-probe frontier under a common policy-visible update budget. Cross-model transport is structural rather than parametric: different model families may use different legal revealability proxies, fitted coefficients, or stopping thresholds, and the theory explicitly allows scalar decision sufficiency whenever no positive-mass scalar fiber crosses the cost-adjusted continuation boundary. Fully tail-robust population utility certification remains conditional on an explicit severity, moment, or tail class.

\paragraph{Organization.}
\Cref{sec:decision-value,sec:revelation-depth} identify the decision quotient and the depth at which hidden distinctions become decision sufficient. \Cref{sec:productive-technology} develops productive fork--probe--promote execution. \Cref{sec:adaptive-depth} closes static refinement into state-dependent control, proves the scalar-control factorization criterion, and gives the model-specific instantiation protocol. \Cref{sec:learnable-control} separates oracle value from finite-sample extraction and certification; the related-work section then fixes the novelty boundary before the empirical instantiation. \Cref{sec:transformer,sec:finite-library-replication,sec:frontier,sec:structural-refinement,sec:frontier-performance,sec:adaptive-validation} evaluate the theory in two Transformer families, and \cref{sec:crossmodel-synthesis} summarizes the cross-model invariants and regime-specific differences. \Cref{sec:applications} then identifies prospective learning, computational, physical, scientific, operational, and high-stakes domains in which the same theory-first instantiation procedure may be useful. Limitations and conclusions follow; the appendices collect proofs, statistical inference, and experimental protocol details.

\section{Decision value and decision factorization}
\label{sec:decision-value}

Let $(\Omega,\mathcal F,\Pp)$ be a probability space, $\mathcal A$ a finite action set, and $Q_a\in L^1$ the terminal utility of action $a$.  For an information sigma-field $\mathcal I\subseteq\mathcal F$, define the observation-relative Bayes value
\begin{equation}
V_B(\mathcal I)
=
\E\!\left[\max_{a\in\mathcal A}\E[Q_a\mid\mathcal I]\right].
\label{eq:bayes-value}
\end{equation}
If $\HB\subseteq\GB$, then Jensen's inequality for the finite maximum gives the standard refinement monotonicity
\begin{equation}
\mathcal I(\GB\mid\HB):=V_B(\GB)-V_B(\HB)\ge0.
\label{eq:refinement-value}
\end{equation}
For an event $A\in\HB$ with $\Pp(A)>0$, define
\begin{equation}
V_B(\mathcal I;A)
=
\E\!\left[\max_{a\in\mathcal A}\E[Q_a\mid\mathcal I]\,\middle|\,A\right],
\label{eq:conditional-bayes-value}
\end{equation}
and for a measurable policy $\pi$, write $\Vpol(\pi;A)=\E[Q_\pi\mid A]$ and $\Vpol(\pi)=\E[Q_\pi]$.

\subsection{Decision factorization}

\begin{definition}[Decision factorization]
For $\HB\subseteq\GB$, the $\GB$-optimal decision \emph{factorizes through} $\HB$ if there exists an $\HB$-measurable policy $\pi_H$ satisfying
\begin{equation}
\pi_H(\omega)\in\arg\max_{a\in\mathcal A}\E[Q_a\mid\GB](\omega)
\quad\text{a.s.}
\end{equation}
The definition concerns the action-relevant quotient of the refined information, not reconstruction of the full latent state.
\end{definition}

\begin{proposition}[Exact finite-action strictness criterion]
\label{prop:factorization-criterion}
For finite $\mathcal A$ and integrable utilities,
\begin{equation}
V_B(\GB)=V_B(\HB)
\end{equation}
if and only if the $\GB$-optimal decision factorizes through $\HB$.  Hence strict refinement value occurs exactly when no $\HB$-measurable policy is $\GB$-Bayes optimal almost surely.
\end{proposition}

This criterion is deliberately decision-relative.  A refinement can contain predictive information while having zero value for the declared action menu; conversely, only a small quotient of a very high-dimensional latent state may be required to change the optimal action.  This perspective is compatible with classical comparison of experiments and value-of-information theory \citep{blackwell1953equivalent,howard1966information,berger1985statistical}, but our later use is dynamic because the act of acquiring future-learning information can also create reusable computation.

\subsection{Dynamic decision-factorization obstruction}

At time $t$, let $\nu_t(a)$ be the current coarse-information action value before accounting for unresolved downstream distinctions, and let $\nu_t^\star=\max_a\nu_t(a)$.  Define the current opportunity cost
\begin{equation}
d_t(a)=\nu_t^\star-\nu_t(a)\ge0.
\end{equation}
Let $L_t\ge0$ be the immediate value lost because the coarse representation aliases current decision-relevant states, and let $\omega_t(a)\ge0$ be the continuation obstruction that remains after taking action $a$.  With continuation factor $\gamma\ge0$, define the total dynamic obstruction by comparing the best continuation-aware action with the coarse current benchmark:
\begin{equation}
\mathcal O_t^{\rm DF}
:=
L_t+
\max_{a\in\mathcal A}\{\nu_t(a)+\gamma\omega_t(a)\}
-\nu_t^\star.
\end{equation}

\begin{proposition}[Dynamic decision-factorization decomposition]
\label{prop:df}
The obstruction admits the exact decomposition
\begin{equation}
\boxed{
\mathcal O_t^{\rm DF}
=
L_t+\max_{a\in\mathcal A}\{\gamma\omega_t(a)-d_t(a)\}.
}
\label{eq:df}
\end{equation}
Moreover,
\begin{equation}
\mathcal O_t^{\rm DF}=0
\iff
L_t=0
\quad\text{and}\quad
\gamma\omega_t(a)\le d_t(a)\ \forall a.
\end{equation}
In particular, if $\gamma>0$ and any current coarse-optimal action has $\omega_t(a)>0$, then the coarse representation is dynamically insufficient even when $L_t=0$.
\end{proposition}

The theorem is a decomposition, not a replacement for POMDP or dual-control theory.  It isolates a failure mode useful for learning systems: a summary may support the correct action \emph{now} yet collapse distinctions that become decision-relevant after the chosen action enters a new training state.

\section{Decision-sufficient revelation and revelation depth}
\label{sec:revelation-depth}

A learning execution state is broader than model parameters:
\begin{equation}
X_t=(\theta_t,m_t,v_t,\text{history},\text{RNG},\text{scheduler},\ldots).
\end{equation}
The legal current information is
\begin{equation}
\HB_t=\sigma(C_t),
\end{equation}
where $C_t$ may contain all prospectively legal current losses, readouts, history summaries, optimizer summaries, and telemetry.  ``Current-only'' therefore does not mean behavior-only.  A controlled future-learning probe $Y_h$ induces
\begin{equation}
\GB_h=\HB_t\vee\sigma(Y_h).
\end{equation}

The companion work establishes predictive non-sufficiency but not decision value \citep{wang2026fiber}. The present section asks which parts of a current-behavior fiber must be revealed to factor the terminal decision quotient. For the binary action menu $\{\KEEP,\XI\}$, let $q_a(x)$ denote the state-conditional expected terminal utility of action $a$ under the declared evaluation technology, and define the terminal action gap
\begin{equation}
g(x)=q_{\XI}(x)-q_{\KEEP}(x).
\end{equation}
A pointwise decision-aliasing witness consists of $x_1,x_2$ such that
\begin{equation}
C(x_1)=C(x_2),
\qquad
g(x_1)g(x_2)<0.
\label{eq:pointwise-alias-witness}
\end{equation}
The full state need not be recovered: the probe only needs to refine the \emph{decision quotient}, i.e., the distinctions necessary to select an optimal action.

\begin{definition}[Decision-relevant revelation]
A future-learning probe $Y_h$ is decision-relevant relative to $\HB_t$ if
\begin{equation}
V_B(\GB_h)>V_B(\HB_t).
\end{equation}
\end{definition}

\subsection{Exact binary aliasing identity}

Let $D=Q_{\XI}-Q_{\KEEP}$, $m_H=\E[D\mid\HB]$, and $m_G=\E[D\mid\GB]$.  Define the $\HB$-measurable random variables
\begin{equation}
a_H=\E[(m_G)_+\mid\HB],
\qquad
b_H=\E[(-m_G)_+\mid\HB].
\end{equation}
By the tower property, $m_H=a_H-b_H$ almost surely.

\begin{theorem}[Binary aliasing identity]
\label{thm:alias-cell}
For the binary action menu, the exact refinement value is
\begin{equation}
\boxed{
V_B(\GB)-V_B(\HB)
=
\E[\min\{a_H,b_H\}].
}
\label{eq:alias-cell-gap}
\end{equation}
Thus strict value is present exactly when the coarse information has positive probability of retaining refined posterior mass on both sides of the terminal decision boundary.  More quantitatively, let $A\in\HB$ with $\Pp(A)>0$.  If for some $\delta>0$ and $\alpha,\beta>0$,
\begin{equation}
\Pp(m_G\ge\delta\mid\HB)\ge\alpha,
\qquad
\Pp(m_G\le-\delta\mid\HB)\ge\beta
\qquad\text{a.s.\ on }A,
\label{eq:alias-sign-mass}
\end{equation}
then
\begin{equation}
\boxed{
V_B(\GB)-V_B(\HB)
\ge
\Pp(A)\,\delta\,\min(\alpha,\beta)>0.
}
\label{eq:alias-lower-bound}
\end{equation}
\end{theorem}

Equation~\eqref{eq:alias-cell-gap} is the distributional form of the alias-cell argument: within each coarse-information fiber, refined states can favor opposite terminal actions, and only the smaller of the two conditional magnitude-weighted sign masses creates irreducible coarse decision regret.  Predictive variation that never crosses the declared decision boundary has zero value for this binary action menu.  No positive-probability atom of the coarse information is required.

\subsection{Local geometry and revelation depth}

Let $\Phi_c$ denote the complete legal frontier information map after a short probe depth $c$ (current information plus the frontier summary), let $g$ be the terminal action-gap function defined above, and let $Y_h$ be a deeper revelation observation.

\begin{proposition}[Local aliasing and revelation]
\label{prop:local-alias}
Suppose $\Phi_c$, $g$, and $Y_h$ are continuously differentiable near $x_0$, $g(x_0)=0$, and $\Phi_c$ has locally constant rank.  If there exists
\begin{equation}
v\in\ker D\Phi_c(x_0)
\end{equation}
with
\begin{equation}
Dg(x_0)v\neq0,
\qquad
DY_h(x_0)v\neq0,
\end{equation}
then a local curve contained in the frontier level set passes through states requiring opposite terminal actions, while the deeper observation changes to first order.  The frontier information is therefore locally decision-aliased along $v$, whereas the depth-$h$ probe is locally sensitive to that direction.
\end{proposition}

\begin{definition}[Revelation depth]
For a decision-relevant direction $v$ at $x_0$, define its first revelation depth
\begin{equation}
\tau^\star(v)=\inf\{h:DY_h(x_0)v\neq0\}.
\end{equation}
For a discrete probe grid, the infimum is understood over the admissible depths. The quantity records first sensitivity only; it does not by itself assume that sensitivity must persist at every deeper exact-depth readout.
\end{definition}

\begin{corollary}[Equal-budget short-probe separation]
\label{cor:depth-separation}
Suppose the conditions of Proposition~\ref{prop:local-alias} hold and a restart frontier reaches depth $c$ while a productive active method reaches $h=\frac{m}{m-1}c$ at equal update budget.  If
\begin{equation}
DY_c(x_0)v=0,
\qquad
DY_h(x_0)v\neq0,
\end{equation}
then the short-probe information is locally blind to a decision-changing direction that the deeper active observation reveals. On a discrete admissible depth grid---or more generally when the first sensitive depth is attained---persistence over the admissible depth family (for example because the legal depth-$h$ record retains earlier readouts) makes this condition equivalently summarized by $c<\tau^\star(v)\le h$.  If, in addition, this aliasing occurs on a positive-probability coarse-information region and the refined conditional action gap places nonzero mass on both signs there, Theorem~\ref{thm:alias-cell} gives strict Bayes refinement value.
\end{corollary}

For the prespecified two-action contract, the persistent/cumulative-readout shorthand for the regime of interest is
\begin{equation}
\boxed{4<\tau^\star(v)\le8,}
\end{equation}
while the exact nonpersistent condition is $DY_4(x_0)v=0$ and $DY_8(x_0)v\neq0$. This is a \emph{conditional theory-level separation} from an H4 DynaMiCS-style short-probe information class.  It does not assert that every empirical H4 representation satisfies the antecedent; the Transformer study instead tests whether the fixed H8 observation produces reproducible decision-relevant refinement in the declared frontier-ambiguity regime.

\begin{figure}[tbp]
\centering
\resizebox{0.92\linewidth}{!}{\begin{tikzpicture}[x=1cm,y=1cm]
  \node[ffpanelaccent] at (0.00,4.55) {A\quad Coarse Alias};
  \draw[line width=1.0pt,draw=FFGrayDark] (0.00,4.33)--(3.10,4.33);
  \node[ffdashframe,minimum width=2.55cm,minimum height=2.55cm] (fiber) at (1.55,2.65) {};
  \node[fflabel,text=FFGrayDark] at (1.55,3.66) {$C(x_1)=C(x_2)$};
  \node[ffneutral,minimum width=1.38cm,minimum height=.60cm] (x1) at (1.55,2.90) {$x_1$};
  \node[ffneutral,minimum width=1.38cm,minimum height=.60cm] (x2) at (1.55,1.83) {$x_2$};

  \node[ffpanelaccent] at (3.55,4.55) {B\quad Same Future-Learning Probe};
  \draw[line width=1.0pt,draw=FFGrayDark] (3.55,4.33)--(8.05,4.33);
  \node[ffblue,minimum width=2.90cm,minimum height=.76cm] (y1) at (5.85,2.90) {Response $Y_h(x_1)$};
  \node[ffred,minimum width=2.90cm,minimum height=.76cm] (y2) at (5.85,1.83) {Response $Y_h(x_2)$};
  \draw[ffarrowblue] (x1.east)--node[ffsubtle,above] {$Y_h$} (y1.west);
  \draw[ffarrowred] (x2.east)--node[ffsubtle,below] {$Y_h$} (y2.west);

  \node[ffpanelaccent] at (8.72,4.55) {C\quad Decision Split};
  \draw[line width=1.0pt,draw=FFGrayDark] (8.72,4.33)--(12.35,4.33);
  \node[ffblue,minimum width=2.45cm,minimum height=.72cm] (a1) at (10.35,2.90) {$\KEEP$ Optimal};
  \node[ffred,minimum width=2.45cm,minimum height=.72cm] (a2) at (10.35,1.83) {$\XI$ Optimal};
  \draw[ffarrowblue] (y1)--(a1);
  \draw[ffarrowred] (y2)--(a2);

  \draw[ffrule] (0.00,.78)--(12.35,.78);
  \node[fflabel,text=FFGrayDark,align=center,text width=10.8cm] at (6.18,.28)
    {Revelation has decision value only when a currently aliased distinction is separated across the action boundary.};
\end{tikzpicture}}
\caption{Decision-relative revelation. A coarse information cell can contain learning states that require different terminal actions. Deeper future-learning responses are useful only when they separate that decision quotient; full hidden-state reconstruction is unnecessary.}
\label{fig:aliasing-revelation}
\end{figure}
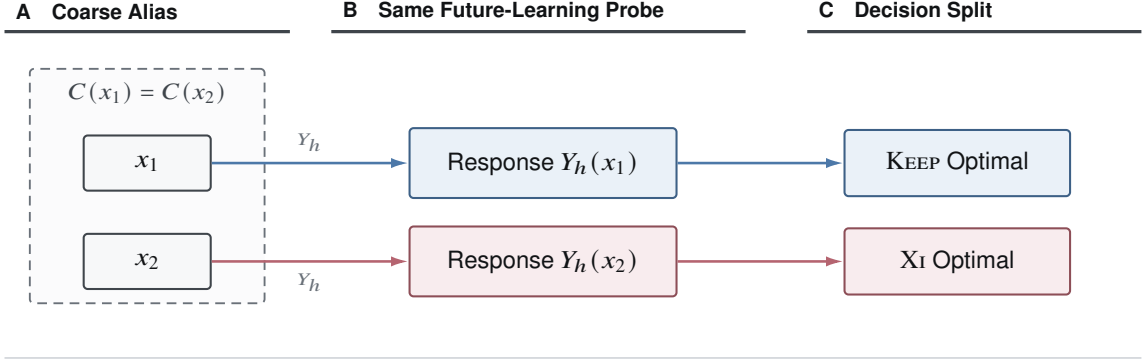

\section{Productive revelation technology}
\label{sec:productive-technology}

Having identified what a useful refinement must reveal, we next ask how that information is acquired and what state the experiment leaves behind. A trial protocol $u$ of depth $h$ produces an observation $Y_{u,h}$ and the refined information
\begin{equation}
\GB_{u,h}=\HB\vee\sigma(Y_{u,h}).
\end{equation}
We distinguish two execution technologies.

\paragraph{Restart / temporary probing.}
The tested path is discarded.  After a decision is made, the selected action starts freshly from the original anchor.

\paragraph{Promotion / productive revelation.}
The selected tested path is retained and continued from its trial endpoint.  The probe is therefore simultaneously an information acquisition and a partially executed candidate action.

The distinction matters because pure information refinement and technology expansion are different objects.  Promotion is not ``free compute'': it is credit for work that remains valid on the selected deployment path.  Conversely, when a probe is unsafe, destructive, or otherwise unusable for deployment, restart is the correct technology and the identity below does not apply.

\subsection{Exact equal-budget identity}

Suppose there are $m\ge2$ candidate actions and a common terminal horizon $H$, with admissible probe depths $c,h\in[0,H]$.  A temporary/restart method that probes every action to depth $c$ and then executes the selected action freshly to $H$ spends
\begin{equation}
K_F(c)=mc+H.
\end{equation}
A probe-and-promote method that probes every action to depth $h$ and then continues the selected tested path spends
\begin{equation}
K_{\AR}(h)=mh+(H-h)=H+(m-1)h.
\end{equation}

\begin{proposition}[Productive revelation equal-budget identity]
\label{prop:productive-budget}
If $K_F(c)=K_{\AR}(h)$, then
\begin{equation}
\boxed{h=\frac{m}{m-1}c.}
\end{equation}
For two actions, $h=2c$.
\end{proposition}
Because $h\le H$, an equal-budget productive match within the declared horizon is feasible only when $c\le (m-1)H/m$; this condition is satisfied by the H4/H8/H12 comparison below.

For the prespecified Transformer comparison, $m=2$, $H=12$, and both methods receive 20 policy-visible updates:
\begin{equation}
K_F(4)=2(4)+12=20,
\qquad
K_{\AR}(8)=12+8=20.
\end{equation}
Thus an H4 temporary probe and an H8 productive probe are compute-matched under this contract.

\begin{figure}[tbp]
\centering
\resizebox{0.96\linewidth}{!}{\begin{tikzpicture}[x=0.50cm,y=0.94cm]
  \node[ffpanelaccent] at (0.00,4.18) {Equal-Budget Compute Geometry};
  \node[ffsubtle,anchor=east,text=FFGrayDark] at (20.00,4.18) {20 policy-visible updates per policy};
  \draw[line width=.82pt,draw=FFRule] (0.00,3.94)--(20.00,3.94);

  \node[fflabel,anchor=east] at (-.62,3.08) {Restart frontier};
  \node[fflabel,anchor=east] at (-.62,1.48) {Productive revelation};

  \fill[FFLight] (0,2.74) rectangle (4,3.42);   \draw[ffline] (0,2.74) rectangle (4,3.42);
  \fill[FFLight] (4,2.74) rectangle (8,3.42);   \draw[ffline] (4,2.74) rectangle (8,3.42);
  \fill[white]   (8,2.74) rectangle (20,3.42); \draw[ffline] (8,2.74) rectangle (20,3.42);
  \node[fflabel] at (2,3.08) {$\KEEP$ probe H4};
  \node[fflabel] at (6,3.08) {$\XI$ probe H4};
  \node[fflabel] at (14,3.08) {Fresh selected trajectory to H12};
  \node[font=\sffamily\scriptsize,text=FFGray] at (4,2.47) {8 probe updates discarded};
  \node[font=\sffamily\scriptsize,text=FFGray] at (14,2.47) {12 fresh execution updates};

  \fill[FFBlueLight] (0,1.14) rectangle (8,1.82);   \draw[draw=FFBlue!80!black,line width=.64pt] (0,1.14) rectangle (8,1.82);
  \fill[FFRedLight]  (8,1.14) rectangle (16,1.82); \draw[draw=FFRed!78!black,line width=.64pt] (8,1.14) rectangle (16,1.82);
  \fill[FFGoldLight] (16,1.14) rectangle (20,1.82);\draw[draw=FFGold!75!black,line width=.64pt] (16,1.14) rectangle (20,1.82);
  \node[fflabel] at (4,1.48) {$\KEEP$ probe H8};
  \node[fflabel] at (12,1.48) {$\XI$ probe H8};
  \node[fflabel] at (18,1.48) {Continue 4};
  \node[font=\sffamily\scriptsize,text=FFGold!78!black] at (8,2.06) {16 probe updates; selected H8 prefix is reusable};

  \node[font=\sffamily\scriptsize\bfseries,text=FFGrayDark,anchor=west] at (20.42,3.08) {$=20$};
  \node[font=\sffamily\scriptsize\bfseries,text=FFGrayDark,anchor=west] at (20.42,1.48) {$=20$};

  \draw[draw=FFGrayDark,line width=.60pt] (0,.37)--(20,.37);
  \foreach \x in {0,4,8,12,16,20}{
    \draw[ffline] (\x,.27)--(\x,.47);
    \node[ffsubtle,anchor=north] at (\x,.18) {\x};
  }
  \node[ffsubtle,anchor=west] at (20.35,.18) {Updates};
\end{tikzpicture}}
\caption{Equal-budget compute geometry for the two-action H12 comparison. Restart probing discards both H4 trials and executes the selected action freshly; productive revelation probes both actions to H8 and retains the selected tested prefix. Both consume 20 policy-visible updates, but promotion buys twice the revelation depth in the two-action case.}
\label{fig:budget-geometry}
\end{figure}
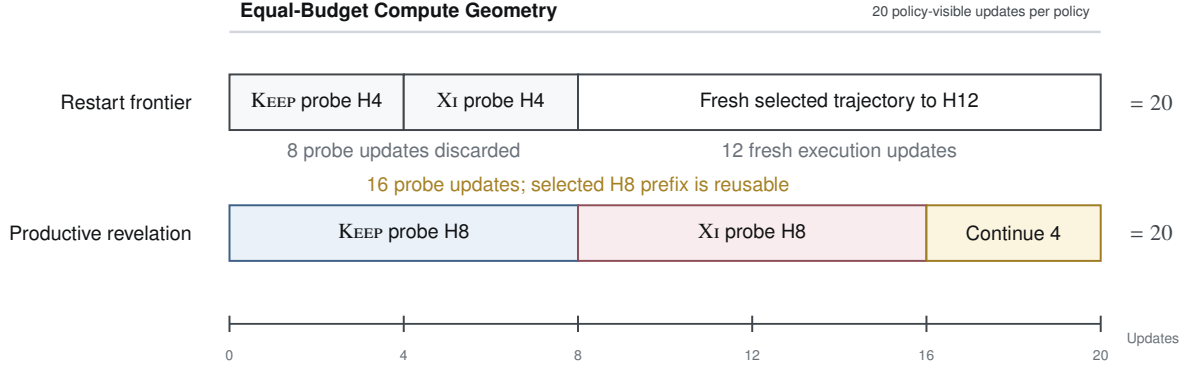

\subsection{End-to-end value decomposition}

For any fixed active policy $\pi_A$ and frontier policy $\pi_F$, let $V_R$ denote common-restart evaluation and $V_P$ productive evaluation of the active policy.  Adding and subtracting $V_R(\pi_A)$ gives the exact identity
\begin{equation}
\boxed{
V_P(\pi_A)-V_R(\pi_F)
=
\underbrace{V_R(\pi_A)-V_R(\pi_F)}_{\Delta_{\rm restart}}
+
\underbrace{V_P(\pi_A)-V_R(\pi_A)}_{\Delta_{\rm path}}.
}
\label{eq:end-to-end-decomp}
\end{equation}
The first term measures policy quality on common restart outcomes; the second measures productive reuse of the tested path.  This identity is especially useful empirically because the two components can be estimated on independent panels without mixing absolute outcomes from unrelated future banks.

\section{Adaptive revelation control: the dynamic closure}
\label{sec:adaptive-depth}

The preceding sections treat $\HB\subseteq\GB$ as a fixed information refinement. Once future-learning depth can be chosen, the information state itself becomes a control variable. This section shows that the dynamic problem is not a separate objective: it is the conditional version of the same Bayes refinement value in Equation~\eqref{eq:refinement-value}. The generic ``continue if expected decision improvement exceeds cost'' principle is classical metareasoning \citep{russell1991principles,frazier2008knowledge}; the learning-state specialization here supplies the revelation filtration, productive cost geometry, and revealability state.

Let $\{\mathscr F_h\}$ be the legal revelation filtration, nested in probe depth $h$, and define the Bayes commit value
\begin{equation}
C_h=\max_{a\in\mathcal A}\E[Q_a\mid\mathscr F_h].
\label{eq:commit-value-depth}
\end{equation}
For $h'>h$, define the oracle conditional value of purchasing the deeper refinement
\begin{equation}
\mathcal G_{h\to h'}=\E[C_{h'}\mid\mathscr F_h]-C_h.
\label{eq:oracle-vor}
\end{equation}

\begin{proposition}[Static-to-dynamic embedding]
\label{prop:static-dynamic-embedding}
For every nested pair $\mathscr F_h\subseteq\mathscr F_{h'}$,
\begin{equation}
\boxed{\E[\mathcal G_{h\to h'}]=V_B(\mathscr F_{h'})-V_B(\mathscr F_h)=\mathcal I(\mathscr F_{h'}\mid\mathscr F_h).}
\label{eq:static-dynamic-embedding}
\end{equation}
Thus the static Bayes refinement value is the population average of the state-dependent continuation value used by adaptive revelation.
\end{proposition}

For fixed shallow and deep decision policies $\pi_h$ and $\pi_{h'}$, define the legal-information policy-level gain
\begin{equation}
G_{h\to h'}=\E\!\left[Q_{\pi_{h'}}-Q_{\pi_h}\,\middle|\,\mathscr F_h\right].
\label{eq:vor-g}
\end{equation}
When the policies are Bayes optimal for their legal information, this coincides with Equation~\eqref{eq:oracle-vor}. With $m$ candidate actions and productive promotion, extending every candidate from $h$ to $h'$ costs only
\begin{equation}
\Delta K(h,h')=(m-1)(h'-h)
\label{eq:adaptive-marginal-compute}
\end{equation}
additional policy-visible updates. At update price $\lambda$, write $c_{h\to h'}=\lambda(m-1)(h'-h)$.

\begin{proposition}[Conditional revelation stopping]
\label{prop:vor-stopping}
Relative to stopping at depth $h$, the pointwise optimal one-step continuation decision for a fixed policy pair is
\begin{equation}
\boxed{\text{continue to }h'\iff G_{h\to h'}>c_{h\to h'}.}
\label{eq:vor-stopping}
\end{equation}
For Bayes policies, the same rule uses $\mathcal G_{h\to h'}$. A population ambiguity quantile is therefore not, in general, a cost-optimal stopping boundary.
\end{proposition}

\begin{theorem}[Cost-adjusted scalar-control factorization]
\label{thm:scalar-control-factorization}
Let $S_h$ be any declared scalar summary measurable with respect to $\mathscr F_h$, and let
\begin{equation}
\Gamma_{h\to h'}:=G_{h\to h'}-c_{h\to h'}
\label{eq:net-continuation-gain}
\end{equation}
be the conditional net gain from continuing rather than stopping.  Relative to always stopping, define the one-step meta-control values
\begin{align}
V_{\rm meta}(\mathscr F_h)
&=\E[(\Gamma_{h\to h'})_+],\\
V_{\rm meta}(S_h)
&=\E\!\left[\left(\E[\Gamma_{h\to h'}\mid S_h]\right)_+\right].
\end{align}
If
\begin{equation}
a(S_h)=\E[(\Gamma_{h\to h'})_+\mid S_h],
\qquad
b(S_h)=\E[(-\Gamma_{h\to h'})_+\mid S_h],
\end{equation}
then the exact value of retaining the full legal shallow information rather than only $S_h$ is
\begin{equation}
\boxed{
V_{\rm meta}(\mathscr F_h)-V_{\rm meta}(S_h)
=
\E[\min\{a(S_h),b(S_h)\}].
}
\label{eq:scalar-control-gap}
\end{equation}
Consequently,
\begin{equation}
V_{\rm meta}(\mathscr F_h)=V_{\rm meta}(S_h)
\end{equation}
if and only if, conditional on $S_h$, the legal shallow states do not place positive probability on both $\Gamma_{h\to h'}>0$ and $\Gamma_{h\to h'}<0$ on any set of positive probability.  Equivalently, there exists an optimal \emph{\textsc{Stop}/\textsc{Continue}} meta-action that factorizes through $S_h$; ties at zero are value-neutral.
\end{theorem}

Theorem~\ref{thm:scalar-control-factorization} sharpens the distinction between predictive and decision nonredundancy.  A second shallow coordinate may change the numerical continuation value while remaining irrelevant to the optimal meta-action if every state within a scalar fiber stays on the same side of the priced stopping boundary.  What makes an additional revealability coordinate \emph{decision-nonredundant} is not conditional variation by itself, but cost-adjusted boundary crossing within a scalar fiber.

\begin{proposition}[Finite-depth Bellman closure]
\label{prop:adaptive-bellman}
Let $h_0<\cdots<h_J$ be admissible revelation depths and let $c_j$ be the priced marginal cost of moving from $h_j$ to $h_{j+1}$. If deeper revelation can be purchased sequentially, then
\begin{align}
V_J^{\rm dyn}&=C_{h_J},\\
V_j^{\rm dyn}&=\max\left\{C_{h_j},\;\E[V_{j+1}^{\rm dyn}\mid\mathscr F_{h_j}]-c_j\right\}.
\label{eq:adaptive-bellman}
\end{align}
Thus static information refinement becomes an optimal-stopping control problem over the revelation filtration. In a general fork--probe--promote system the state must also retain promoted-state information needed to make the transition Markov.
\end{proposition}

\subsection{Margin and revealability: value variation versus decision nonredundancy}

For binary actions, let
\begin{equation}
D=Q_{\XI}-Q_{\KEEP},
\qquad
m_h=\E[D\mid\mathscr F_h].
\end{equation}
Suppose the refined Bayes margin admits the conditional location--scale representation
\begin{equation}
m_{h'}=m_h+\sigma_h Z,
\label{eq:location-scale-refinement}
\end{equation}
where $Z$ is independent of $\mathscr F_h$, symmetric about zero, integrable, and nondegenerate, while $\sigma_h\ge0$ is $\mathscr F_h$-measurable. Define
\begin{equation}
\psi(t)=\E[(Z-t)_+],\qquad t\ge0,
\end{equation}
and define the conditional Bayes value of deeper revelation by
\begin{equation}
\mathcal R_{h\to h'}
:=
\E[(m_{h'})_+\mid\mathscr F_h]-(m_h)_+.
\label{eq:conditional-revelation-value}
\end{equation}

\begin{theorem}[Two-coordinate revelation geometry]
\label{thm:two-coordinate-revelation}
Under Equation~\eqref{eq:location-scale-refinement}, the conditional Bayes value of deeper revelation is
\begin{equation}
\boxed{\mathcal R_{h\to h'}=\sigma_h\psi\!\left(\frac{|m_h|}{\sigma_h}\right),}
\label{eq:two-coordinate-value}
\end{equation}
with the convention $\mathcal R_{h\to h'}=0$ when $\sigma_h=0$. Wherever differentiable,
\begin{equation}
\frac{\partial\mathcal R}{\partial |m_h|}
=-\Pr\!\left(Z>\frac{|m_h|}{\sigma_h}\right)\le0,
\end{equation}
and
\begin{equation}
\frac{\partial\mathcal R}{\partial\sigma_h}
=\psi(t)+t\Pr(Z>t)\ge0,
\qquad t=\frac{|m_h|}{\sigma_h}.
\end{equation}
Hence, in the nonredundant regime, the cost-optimal continuation boundary is a curve in
\begin{equation}
\boxed{(|m_h|,\sigma_h),}
\end{equation}
rather than a universal threshold on current decision margin. The derivative in $\sigma_h$ is strict whenever the refinement retains positive probability mass beyond the current normalized margin. The theorem itself does not require $\sigma_h$ to remain empirically nonredundant after conditioning on $|m_h|$.
\end{theorem}

\begin{corollary}[Conditional value nonredundancy and scalar reduction]
\label{cor:conditional-revealability}
Let $M_h=|m_h|$ and write
\begin{equation}
r(m,s)=s\,\psi(m/s),
\end{equation}
with the same zero-scale convention as in Theorem~\ref{thm:two-coordinate-revelation}.
\begin{enumerate}[label=(\roman*)]
\item \emph{Scalar-redundant regime.} If $\sigma_h=g(M_h)$ almost surely on a decision-relevant event $E$ for some measurable $g$, then
\begin{equation}
\mathcal R_{h\to h'}=\widetilde r(M_h)
\qquad\text{on }E,
\end{equation}
for the scalar function $\widetilde r(m)=r(m,g(m))$. Hence current margin is sufficient for the one-step oracle continuation value on $E$. A single monotone margin threshold requires the additional condition that $\widetilde r$ be monotone relative to the fixed compute price.
\item \emph{Nonredundant regime.} If the conditional law of $\sigma_h$ given $M_h$ is nondegenerate on a set of positive probability and, on the corresponding conditional support, $s\mapsto r(M_h,s)$ is strictly increasing, then the conditional law of $\mathcal R_{h\to h'}$ given $M_h$ is also nondegenerate on a set of positive probability. Consequently no measurable function of current margin alone can reproduce the oracle continuation value there.
\end{enumerate}
\end{corollary}

\begin{corollary}[Cost-adjusted revealability crossing criterion]
\label{cor:cost-adjusted-revealability-crossing}
Assume the setting of Theorem~\ref{thm:two-coordinate-revelation}, fix a continuation price $c>0$, and suppose that for almost every $m$ in the decision-relevant region the map $s\mapsto r(m,s)$ is continuous and strictly increasing on the conditional support of $\sigma_h\mid M_h=m$. Define the generalized critical revealability
\begin{equation}
\sigma_c(m)=\inf\{s\ge0:r(m,s)>c\},
\label{eq:critical-revealability}
\end{equation}
with $\inf\varnothing=+\infty$.
\begin{enumerate}[label=(\roman*)]
\item The scalar-margin decision is sufficient if and only if, for almost every $M_h$, either
\begin{equation}
\Pr\!\left(r(M_h,\sigma_h)\le c\mid M_h\right)=1
\quad\text{or}\quad
\Pr\!\left(r(M_h,\sigma_h)\ge c\mid M_h\right)=1.
\label{eq:crossing-sufficiency}
\end{equation}
Equivalently, conditional on almost every margin value, the priced continuation gain has no mass on both strict sides of zero. Under the stated continuity and monotonicity conditions, a transparent sufficient support condition is that $\sigma_h\mid M_h=m$ is contained in $[0,\sigma_c(m)]$ or in $[\sigma_c(m),\infty)$; zero-gain ties at the boundary are value-neutral and may be assigned to either optimal meta-action.
\item If there is a set of margin values with positive probability on which
\begin{equation}
\Pr\!\left(r(M_h,\sigma_h)<c\mid M_h\right)>0
\quad\text{and}\quad
\Pr\!\left(r(M_h,\sigma_h)>c\mid M_h\right)>0,
\label{eq:revealability-crossing}
\end{equation}
then the scalar-margin controller is strictly suboptimal:
\begin{equation}
V_{\rm meta}(\mathscr F_h)>V_{\rm meta}(M_h).
\end{equation}
Under the same monotonicity conditions, Equation~\eqref{eq:revealability-crossing} is implied by positive conditional mass strictly below and strictly above $\sigma_c(M_h)$. Thus conditional spread in revealability is \emph{decision-nonredundant} exactly when a positive-mass scalar fiber contains both the strict \textsc{Stop} side and the strict \textsc{Continue} side of the priced continuation decision; zero-gain ties at the boundary do not affect value.
\end{enumerate}
\end{corollary}

\begin{proposition}[Location--scale identification of revealability scale]
\label{prop:revealability-scale-identification}
Under Equation~\eqref{eq:location-scale-refinement},
\begin{equation}
\E\!\left[|m_{h'}-m_h|\mid\mathscr F_h\right]
=\sigma_h\,\E|Z|.
\label{eq:abs-scale-identification}
\end{equation}
If additionally $\E[Z^2]=1$, then
\begin{equation}
\E\!\left[(m_{h'}-m_h)^2\mid\mathscr F_h\right]=\sigma_h^2.
\label{eq:sq-scale-identification}
\end{equation}
At the population level, these identities identify revealability scale up to a fixed normalization from the conditional spread of the true deeper-margin increment. In applications the Bayes margins themselves may be latent or estimated; the identities then motivate development-only proxy construction rather than claiming direct identification from noisy fitted scores.
\end{proposition}

\begin{definition}[Legal model-specific revealability proxy]
\label{def:model-specific-revealability-proxy}
For a model family $\mathcal M$, let $X_h^{(\mathcal M)}$ denote a declared shallow information map measurable with respect to $\mathscr F_h$. A statistic
\begin{equation}
R_h^{(\mathcal M)}=r_{\mathcal M}\!\left(X_h^{(\mathcal M)}\right)
\end{equation}
is a \emph{legal model-specific revealability proxy} when its functional form or fitted parameters are determined without using the target outcomes and are fixed before target evaluation. The proxy may estimate $\sigma_h$ directly, rank the conditional spread in Proposition~\ref{prop:revealability-scale-identification}, or enter a predeclared estimator of $G_{h\to h'}$. No equality $R_h^{(\mathcal M)}=\sigma_h$ is assumed, and different model families need not share the same proxy formula, scale, coefficient, or stopping threshold.
\end{definition}

\begin{protocol}[Model-specific Revelation-Control instantiation]
\label{prot:model-specific-instantiation}
For a new learning-system family $\mathcal M$, an empirical instantiation should proceed as follows.
\begin{enumerate}[label=\textbf{P\arabic*.},leftmargin=*,itemsep=2pt]
\item \textbf{Declare the decision problem.} Fix the action menu, terminal utility, legal shallow filtration, admissible revelation depths, productive-promotion technology, and resource-price ledger before target outcomes are inspected.
\item \textbf{Separate development from target evaluation.} Collect a development panel with the shallow and deeper potential outcomes needed to estimate continuation value. Keep the target panel disjoint at the exact-anchor level.
\item \textbf{Instantiate the observable continuation state.} Fit either $G_{h\to h'}$ directly from legal shallow information or a legal proxy $R_h^{(\mathcal M)}$ motivated by Proposition~\ref{prop:revealability-scale-identification}. Always retain a declared scalar comparator.
\item \textbf{Control representation selection.} Choose a minimal learnable shallow summary using development data only. If a finite set of architectures remains eligible, predeclare that family and use simultaneous or familywise-valid inference on the target panel rather than selecting an unadjusted winner after target evaluation.
\item \textbf{Fix the controller before evaluation.} Fix the fitted estimator, stop/continue threshold, tie and fallback rules, productive compute accounting, risk target, resampling unit, and inferential criteria before evaluating target outcomes.
\item \textbf{Evaluate once and separate claims.} Evaluate the fixed family on the disjoint target panel. Report decision value, compute-accounted utility, and decision-instability risk as distinct objects; a bounded-risk claim becomes an expected-utility certificate only under an explicit severity or tail bridge.
\end{enumerate}
The protocol transports the theoretical structure, not the numerical proxy, coefficient, or threshold of an earlier model family.
\end{protocol}

\paragraph{Cross-model transport is structural, not parametric.}
The theorem-level transport target is the Revelation-Control relation between legal shallow information, continuation value, productive cost, and the stopping decision. A second model family need not reproduce the same empirical proxy used in the first. To claim a \emph{decision-nonredundant revealability regime}, a legal proxy specified independently of target outcomes must add held-out control value beyond a scalar-margin comparator, or otherwise establish cost-adjusted boundary crossing as in Corollary~\ref{cor:cost-adjusted-revealability-crossing}. If the model fit on development data and fixed before target evaluation instead admits a successful scalar continuation rule and no legal second coordinate shows incremental control value, that outcome is consistent with scalar decision sufficiency under Theorem~\ref{thm:scalar-control-factorization}, without requiring the stronger assertion that revealability is functionally determined by margin. In either regime, target evaluation must use the same declared terminal utility and compute ledger, and the proxy must not be selected from target outcomes.

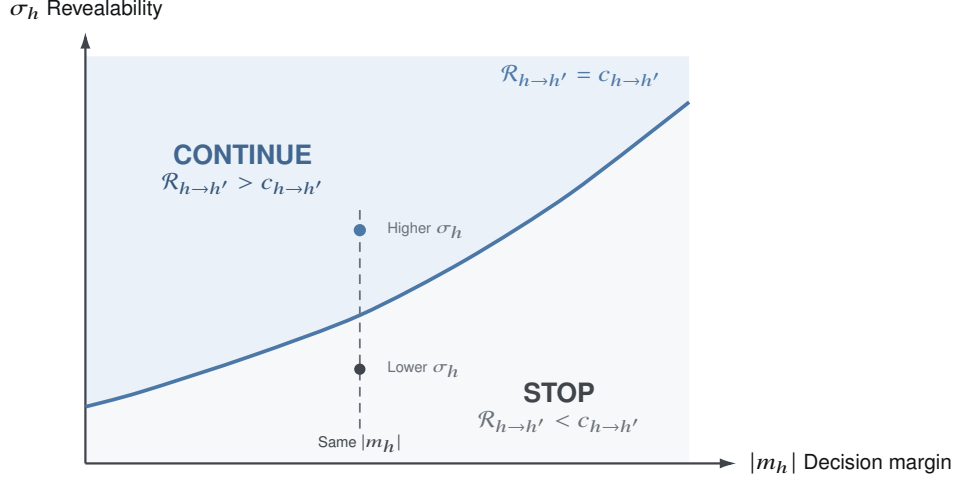
\begin{figure}[tbp]
\centering
\resizebox{0.78\linewidth}{!}{\begin{tikzpicture}[x=1cm,y=1cm]
  \path[fill=FFLight]
    (0,0) -- (6.60,0) -- (6.60,3.95) --
    plot[smooth] coordinates {(5.30,2.95) (4.15,2.22) (3.00,1.62) (1.80,1.17) (.60,.78) (0,.62)} -- cycle;
  \path[fill=FFBlueLight]
    (0,.62) --
    plot[smooth] coordinates {(.60,.78) (1.80,1.17) (3.00,1.62) (4.15,2.22) (5.30,2.95) (6.60,3.95)} --
    (6.60,4.45) -- (0,4.45) -- cycle;

  \draw[-{Latex[length=1.85mm,width=1.12mm]},draw=FFGrayDark,line width=.70pt]
    (0,0)--(7.12,0) node[fflabel,anchor=west] {$|m_h|$  Decision margin};
  \draw[-{Latex[length=1.85mm,width=1.12mm]},draw=FFGrayDark,line width=.70pt]
    (0,0)--(0,4.72) node[fflabel,anchor=south] {$\sigma_h$  Revealability};
  \draw[draw=FFBlue,line width=1.15pt]
    plot[smooth] coordinates {(0,.62) (.60,.78) (1.80,1.17) (3.00,1.62) (4.15,2.22) (5.30,2.95) (6.60,3.95)};
  \node[fflabel,text=FFBlue,anchor=east] at (6.43,4.24) {$\mathcal R_{h\to h'}=c_{h\to h'}$};

  \node[font=\sffamily\small\bfseries,text=FFBlue!85!black] at (1.72,3.38) {CONTINUE};
  \node[font=\sffamily\scriptsize,text=FFBlue!72!black] at (1.72,3.06) {$\mathcal R_{h\to h'}>c_{h\to h'}$};
  \node[font=\sffamily\small\bfseries,text=FFGrayDark] at (5.18,.78) {STOP};
  \node[font=\sffamily\scriptsize,text=FFGray] at (5.18,.46) {$\mathcal R_{h\to h'}<c_{h\to h'}$};

  \draw[ffdashline] (3.00,.38)--(3.00,2.82);
  \fill[FFGrayDark] (3.00,1.03) circle[radius=1.75pt];
  \fill[FFBlue] (3.00,2.55) circle[radius=1.9pt];
  \node[ffsubtle,anchor=west] at (3.18,2.55) {Higher $\sigma_h$};
  \node[ffsubtle,anchor=west] at (3.18,1.03) {Lower $\sigma_h$};
  \node[ffsubtle,text=FFGrayDark] at (3.00,.22) {Same $|m_h|$};
\end{tikzpicture}}
\caption{Schematic margin--revealability geometry implied by Theorem~\ref{thm:two-coordinate-revelation}. At fixed compute price, larger current decision margins require greater state-dependent revealability to justify deeper future learning. The paired points illustrate the nonredundant case in which two states with the same current margin can fall on opposite sides of the stopping boundary because their revealability differs. The curve $\sigma_h=g(|m_h|)$ illustrates the stronger value-redundant special case in which continuation value itself reduces to a scalar function. More generally, scalar \emph{decision} sufficiency requires only that the accessible states at a fixed margin remain on one side of the priced boundary; residual revealability variation is allowed. The boundary is theoretical and schematic, not an empirical fit.}
\label{fig:adaptive-geometry}
\end{figure}

The theorem explains why current confidence and value of further revelation need not induce the same ordering in the nonredundant regime. A state can look locally decisive yet remain highly revealable under future training; conversely, a state near the current decision boundary can have little continuation value if its future response is stable. Corollary~\ref{cor:conditional-revealability} characterizes value-level redundancy, while Theorem~\ref{thm:scalar-control-factorization} and Corollary~\ref{cor:cost-adjusted-revealability-crossing} give the sharper decision-level statement: scalar utility-aware control can remain optimal even with residual revealability variation, provided no positive-mass scalar fiber crosses the priced \textsc{Stop}/\textsc{Continue} boundary.

\section{From oracle revelation to learnable and certifiable control}
\label{sec:learnable-control}

A refinement can improve oracle Bayes value and still be a poor learned representation at finite sample size.  To make this distinction explicit, fix a candidate acquisition scheme $s$ with refined information $\GB_s$ and promotion technology $P_s$.  Let
\begin{align}
\Delta_s^{\rm rev} &= V_B(\GB_s)-V_B(\HB),\\
\Delta_s^{\rm prom} &= V_{P_s}^{\star}-V_B(\GB_s),
\end{align}
where $V_{P_s}^{\star}$ is the best value available under the refined information and the declared promotion technology.  When restart remains an admissible fallback under that technology, $\Delta_s^{\rm prom}\ge0$; otherwise the ledger remains algebraically valid with a signed technology term.  For a learner class $\Pi_s$, let $R_s^{\rm app}\ge0$ be the gap between $V_{P_s}^{\star}$ and the best policy in $\Pi_s$, let $R_{s,n}^{\rm est}\ge0$ be the finite-sample gap between that class optimum and the learned policy, and let $C_s\ge0$ be acquisition cost in the chosen utility units.

\begin{proposition}[Finite-sample revelation ledger]
\label{prop:learnability-ledger}
With the definitions above, the learned net value satisfies the exact identity
\begin{equation}
\boxed{
W_{s,n}^{\rm net}-V_B(\HB)
=
\Delta_s^{\rm rev}
+
\Delta_s^{\rm prom}
-
R_s^{\rm app}
-
R_{s,n}^{\rm est}
-
C_s.
}
\label{eq:learnability-ledger}
\end{equation}
\end{proposition}

The ledger separates an oracle question from an extraction question.  More telemetry can increase $\Delta_s^{\rm rev}$ while increasing effective estimation complexity enough to worsen $R_{s,n}^{\rm est}$.  Development analyses with substantially broader telemetry exhibited this estimation obstruction, motivating a compact refinement rather than maximizing raw feature count.

\subsection{Minimal learnable refinement}

Define the expected finite-sample value
\begin{equation}
\Psi_n(s)
=
\Delta_s^{\rm rev}
+
\Delta_s^{\rm prom}
-
R_s^{\rm app}
-
\E[R_{s,n}^{\rm est}]
-
C_s.
\end{equation}
Let $\kappa_n(s)$ be a learner-relative complexity measure and, for tolerance $\tau\ge0$, let
\begin{equation}
\mathcal S_{n,\tau}
=
\left\{s:\Psi_n(s)\ge\sup_{s'}\Psi_n(s')-\tau\right\}.
\end{equation}
When $\mathcal S_{n,\tau}$ is nonempty and the minimum of $\kappa_n$ is attained on it, a \emph{minimal learnable refinement} is any
\begin{equation}
s_{n,\tau}^{\star}
\in
\arg\min_{s\in\mathcal S_{n,\tau}}\kappa_n(s).
\end{equation}
Minimality is therefore relative to the learner, sample size, acquisition technology, and tolerance.  It need not mean the fewest raw features or the mathematically coarsest sufficient sigma-field.

\subsection{Learned frontier-victory ledger}

Let a fixed frontier learner $\hat\pi_F$ use information $\HB$ under restart evaluation, and let a fixed active learner $\hat\pi_A$ use $\GB\supseteq\HB$.  Define
\begin{equation}
R_F=V_B(\HB)-V_R(\hat\pi_F),
\qquad
R_A=V_B(\GB)-V_R(\hat\pi_A),
\end{equation}
where $V_R$ is common restart-evaluated value, and let $P_A=V_P(\hat\pi_A)-V_R(\hat\pi_A)$ be the productive promotion value of the active policy.

\begin{corollary}[Learned frontier victory]
\label{cor:frontier-ledger}
Writing $\Delta_{\rm rev}=V_B(\GB)-V_B(\HB)$,
\begin{equation}
\boxed{
V_P(\hat\pi_A)-V_R(\hat\pi_F)
=
\Delta_{\rm rev}+P_A+R_F-R_A.
}
\label{eq:frontier-ledger}
\end{equation}
If resources $K_A,K_F$ are priced by $\lambda$, then net value satisfies
\begin{equation}
N_A-N_F
=
\Delta_{\rm rev}+P_A+R_F-R_A
+\lambda^\top(K_F-K_A).
\end{equation}
At equal resource cost, a sufficient condition for active victory is $\Delta_{\rm rev}+P_A>R_A-R_F$.
\end{corollary}

Equation~\eqref{eq:frontier-ledger} separates the two empirical questions studied later: whether deeper future learning reveals decision-relevant structure, and whether productive reuse converts that structure into superior end-to-end utility.

\subsection{Certification boundary for adaptive depth}

Finite-sample risk calibration for early stopping is established prior art: selective prediction, Learn-then-Test, Conformal Risk Control, early-time classification, and risk-controlled early-exit networks provide general mechanisms for calibrating bounded stopping risks \citep{franc2019uncertainty,angelopoulos2025ltt,angelopoulos2024crc,ringel2024earlytime,jazbec2024fast,wang2026conformal}. Productive-prefix reuse exposes a particularly natural structural loss here. If
\begin{equation}
F_{h\to h'}=\ind\{\pi_h\neq\pi_{h'}\},
\end{equation}
then, under the declared same-path promotion technology,
\begin{equation}
F_{h\to h'}=0\quad\Longrightarrow\quad Q_{\pi_h}=Q_{\pi_{h'}}.
\label{eq:decision-invariance-main}
\end{equation}
Thus shallow/deep decision instability can be calibrated with a bounded Bernoulli loss even when terminal utility gaps are heavy-tailed. Turning that probability statement into a strict expected-net-utility guarantee additionally requires control of the stopped-harm tail. Appendix~\ref{app:adaptive-cert} makes this boundary sharp: for any nonzero stop--flip risk, unrestricted severity makes the worst-case expected net utility unbounded below even if stop and flip probabilities are known exactly. Bounded-severity, moment, or integrable-tail assumptions provide sufficient bridges from bounded risk to expected utility.

\section{Related work and novelty boundary}\label{sec:related-work}

\paragraph{Bayesian decision theory and information value.}
The monotonic value of information under refinement is classical in statistical decision theory and comparison of experiments \citep{blackwell1953equivalent,berger1985statistical,howard1966information}.  Expected information from experiments and Bayesian experimental design provide a complementary tradition in which observations are deliberately acquired under a utility criterion \citep{lindley1956measure,chaloner1995bayesian}.  Our Bayes-value notation is a specialization of these traditions, not a new notion of information value or experimental design.  The distinctive question is which \emph{learning-state} distinctions are worth purchasing through a future training intervention once finite-sample extraction and execution technology are included.

\paragraph{Terminology: revelation in decision analysis and mechanism design.}
The phrase \emph{value of revelation} has prior use in influence-diagram decision analysis: \citet{ezawa1998evidence} defines it from values of evidence and relates it to value of control. We therefore do not claim that phrase as new; our conditional quantity in Equation~\eqref{eq:oracle-vor} is a continuation value for an intervention-generated refinement that can also change the controlled system and create reusable progress. The word ``revelation'' also has a different established meaning in mechanism design, where the revelation principle concerns truthful direct mechanisms under private information \citep{myerson1979incentive}. Revelation Control does not assume strategic reporting or incentive compatibility. Its object is controlled acquisition of decision-relevant state through system response.

\paragraph{Partial observability, active sensing, and dual control.}
POMDPs represent decisions under partial observability through belief states \citep{kaelbling1998planning}; dual control emphasizes that an action can simultaneously control a system and reveal uncertainty \citep{feldbaum1960dual}; and active-sensing work studies task-directed information acquisition \citep{veiga2023active}.  Revelation Control belongs to this broad family but does not replace it.  Our narrower object is an internal learning execution state, the intervention is future training, and the formal target is decision factorization through a declared legal information sigma-field.  The promotion term further distinguishes information gained by a trial from useful state change left by that trial.

\paragraph{Predictive state and decision-focused representations.}
Predictive State Representations encode state through action-conditional predictions of future observations \citep{littman2001predictive}.  The companion work is related in spirit because future responses reveal hidden learning-state distinctions, but the present paper does not propose a general predictive state representation.  It asks only for the quotient needed to choose among a fixed action menu.  This has conceptual kinship with state abstraction, where representations are judged by whether they retain distinctions needed for planning or learning \citep{li2006state}. Recent work on decision-relevant concept selection makes this connection explicit by requiring states that share a selected concept representation to preserve the optimal decision structure \citep{raman2026decision}. Decision-focused learning and predict-then-optimize methods likewise emphasize that predictive quality should be judged by downstream decision loss \citep{donti2017task,elmachtoub2022smart,mandi2024decision}; our finite-sample ledger adds an intervention/acquisition dimension in which the representation itself must be actively produced.

\paragraph{Function-equivalent states and self-intervention.}
Recent work on path-conditioned training makes especially clear that ReLU parameterizations implementing the same function can nevertheless induce different subsequent training dynamics \citep{lebeurrier2026path}. This is closely aligned with the motivating non-sufficiency phenomenon, but it studies rescaling and conditioning rather than decision-relative future-learning experiments. Self-Interventional Learning perturbs a network's own functional organization, learns a predictive self-model from intervention consequences, and uses that model for later structural action \citep{tomaszewski2026self}. Revelation Control instead treats the learner as the controlled object of an external decision problem, asks which future-training responses refine a declared action quotient, and accounts separately for information and reusable tested computation.

\paragraph{Partial training, early stopping, and resource allocation.}
Freeze--Thaw Bayesian optimization uses partial learning curves to pause and resume candidate models \citep{swersky2014freeze}; learning-curve extrapolation uses early trajectory segments to predict eventual performance and stop unpromising runs \citep{domhan2015speeding}; and non-stochastic best-arm allocation formalizes the use of intermediate learning performance to concentrate resources on promising configurations \citep{jamieson2016nonstochastic}.  Hyperband extends this resource-allocation perspective at scale \citep{li2018hyperband}, while Population Based Training jointly adapts model populations and hyperparameters while reusing live trajectories \citep{jaderberg2017pbt}.  These works establish strong precedents for partial training, continuation, early termination, promotion, and resource allocation.  Our novelty therefore does not rest on any of those operations individually.

\paragraph{DynaMiCS and local prospective probes.}
Within the declared fine-tuning-probe problem class, DynaMiCS is the closest direct comparator because it explicitly performs short domain-specific fine-tuning probes to estimate local cross-domain slopes before optimizing a constrained data mixture \citep{gualdoni2026dynamics}.  We therefore concede short prospective training probes and local finite-difference summaries as prior art.  Our theory asks a different question: can a local-probe information state fail to factorize the terminal decision, and can productive reuse make a deeper decision-relevant observation available under the same update budget?  The prespecified H4 comparator is a strengthened task adaptation of that acquisition geometry, not a claim that the original DynaMiCS algorithm is globally dominated.

\paragraph{Metareasoning and value of computation.}
Rational metareasoning asks whether an additional computation is worth its cost because of the external decision it may change \citep{russell1991principles}; knowledge-gradient policies similarly choose measurements by expected increment in terminal value \citep{frazier2008knowledge}. Recent adaptive-reasoning work operationalizes marginal-gain-versus-cost allocation through difficulty signals \citep{wu2026coda}, while consequence-aware allocation shows that difficulty and error severity need not coincide when assigning test-time reasoning budgets \citep{wen2026consequence}. Proposition~\ref{prop:vor-stopping} is a Revelation-Control specialization of value-of-computation reasoning, not a new generic stopping rule. The structural point specific to this paper is that Equation~\eqref{eq:static-dynamic-embedding} makes the original hidden-learning-state refinement value the state-dependent reward of that control problem. The distinctive object is hidden learner execution state: the computation is controlled future training, and the selected experimental path can remain useful training.

\paragraph{Risk-controlled stopping.}
Reject-option and selective-classification methods trade prediction coverage against conditional risk \citep{franc2019uncertainty}. Learn-then-Test and Conformal Risk Control provide finite-sample calibration of bounded risks for families of predictive rules \citep{angelopoulos2025ltt,angelopoulos2024crc}. These tools have also been applied directly to sequential stopping: early-time classification can be calibrated for accuracy-gap control \citep{ringel2024earlytime}, risk-controlled neural early exits can use supervised or consistency losses \citep{jazbec2024fast}, and recent work controls reasoning risk under a compute budget \citep{wang2026conformal}. We therefore do not claim to introduce risk-controlled early stopping or early-versus-deep consistency. Our narrower contribution is to connect these ideas to productive future-training probes of hidden learning state, where deeper computation changes the learner and reveals a decision quotient.

\paragraph{Value of computation and finite-sample extraction.}
Our acquisition-cost and learner-regret terms also interact with finite-sample extraction: richer future telemetry can have higher oracle information value and lower learned value because estimation error grows. The minimal-learnable-refinement principle therefore treats representation complexity as part of the decision problem rather than assuming that more legal telemetry is automatically better.

\begin{table}[H]
\centering
\caption{Novelty boundary relative to representative neighboring methods.  Entries describe the central mechanism emphasized by each work, not every implementation variant.}
\label{tab:related-work}
\small
\begin{tabularx}{\linewidth}{>{\raggedright\arraybackslash}p{0.20\linewidth}YYYYY}
\toprule
Method / theory & Partial training & Prospective training signal & Path reuse & Hidden learning-state decision aliasing & Explicit finite-sample value ledger\\
\midrule
Freeze--Thaw \citep{swersky2014freeze} & yes & learning curves & pause/resume & no & no\\
Hyperband \citep{li2018hyperband} & yes & intermediate performance & continuation of survivors & no & no\\
PBT \citep{jaderberg2017pbt} & yes & population performance & yes & no & no\\
DynaMiCS \citep{gualdoni2026dynamics} & yes & local cross-domain slopes & no & no & no\\
Risk-controlled early exit \citep{ringel2024earlytime,jazbec2024fast} & no & shallow/deep prediction signal & no & no & risk control, not utility ledger\\
Revelation Control (this work) & yes & future-learning response & promotion separated from information value & central target & yes\\
\bottomrule
\end{tabularx}
\end{table}

\paragraph{Novelty claim.}
The paper therefore does \emph{not} claim to invent future probing, promotion, Bayes value, value of computation, selective early exit, finite-sample risk calibration, or dynamic control. The strongest claimed contribution is the coupling
\begin{equation}
\boxed{
\begin{gathered}
\text{hidden learning-state aliasing}
+\text{productive future-training experiment}\\
{}+\text{budget-indexed revelation depth}
+\text{finite-sample extraction}
+\text{comparative utility}.
\end{gathered}}
\end{equation}
The adaptive-control layer completes this chain by identifying current decision margin and learning-state revealability as joint control coordinates for purchasing deeper revelation, while also characterizing when revealability is conditionally redundant and scalar control suffices; the generic metareasoning and risk-calibration tools used around that result are explicitly treated as prior art.

\section{Transformer instantiation}
\label{sec:transformer}

We evaluate Revelation Control in two independently instantiated decoder-only 7B Transformer families: Qwen2.5-7B \citep{qwen2024technical,qwen25modelcard} and Mistral-7B-v0.3 \citep{jiang2023mistral,mistralv03modelcard}. Both use LoRA adaptation \citep{hu2022lora} with AdamW optimizer state \citep{kingma2015adam,loshchilov2019adamw}, while the Mistral instantiation uses independently prepared task streams, training histories, and development data. The cross-model target is therefore structural rather than numerical.

In both families, exact anchors are generated from controlled training histories and the binary action menu is
\begin{equation}
\mathcal A=\{\KEEP,\XI\},
\end{equation}
where $\XI$ denotes the prespecified matched optimizer-state intervention. The intervention matches the immediate adaptive field while altering hidden optimizer moments, allowing later common training to reveal a difference that is absent from the immediate update. Terminal utility is evaluated at H12. The active policy probes both candidate actions to H8, selects an action using a decision rule fit on development data and fixed before target evaluation, and promotes the selected tested path to H12. The strengthened short-probe frontier probes to H4, restores the anchor, selects an action from legal H4 information, and restarts that action to H12.

\subsection{Qwen2.5-7B prespecified active policy}

For Qwen2.5-7B, the final fixed-depth active policy is
\begin{equation}
\boxed{
\begin{gathered}
\text{24-dimensional raw H8 response}
+\operatorname{Ridge}(\alpha=10)\\[-1pt]
+\text{zero threshold}
+\text{same-path promotion}.
\end{gathered}
}
\end{equation}
Ties and nonfinite scores fall back to $\KEEP$. The representation and decision rule were selected on development data and fixed before the two independent finite-library panels and the 432-anchor comparative panel.

\paragraph{Selection status.}
The method is prespecified, not a proof of global algorithmic optimality. Development-only comparisons evaluated H2, H4, H6, and H8; H8 was selected before target evaluation, while substantially broader telemetry increased effective estimation burden without a reliable compensating gain.

\subsection{Independent Mistral-7B-v0.3 instantiation}

For Mistral-7B-v0.3, the same H4/H8/H12 decision geometry, action menu, terminal utility, productive-promotion technology, and 20-update accounting are retained, while numerical heads and adaptive summaries are instantiated from Mistral-specific development data. The theory does not require Qwen's empirical revealability proxy, coefficients, or stopping threshold to transfer numerically. A separate 336-anchor development panel supplies the Mistral parameters, and an independent 480-anchor two-bank panel supplies target evaluation. The adaptive analysis is restricted to the scalar and two-coordinate continuation architectures already defined in the Qwen analysis, with Mistral parameters fit only on the development panel. The transported object is the Revelation-Control structure, not one fitted coordinate system. Statistical units, lower bounds, multiplicity control, and bank aggregation are specified in Appendix~\ref{app:inference}.

\section{Fixed-depth revelation beyond current-information comparators}
\label{sec:finite-library-replication}

\begin{figure}[H]
\centering
\resizebox{0.95\linewidth}{!}{\begin{tikzpicture}[font=\tiny,x=1cm,y=.22cm]
  \def\scaleX{0.72}
  \def\panelB{4.60}
  \def\xmax{5.55}

  \fill[FFLight] (-2.55,20.45) rectangle (8.72,22.55);
  \fill[FFLight] (-2.55,10.45) rectangle (8.72,15.55);
  \fill[FFLight] (-2.55,0.45) rectangle (8.72,5.55);

  \node[anchor=west,font=\scriptsize,text=FFGrayDark] at (-2.50,24.05)
    {Open square: simultaneous LCB \quad $\longrightarrow$ \quad filled circle: point estimate};
  \node[font=\scriptsize\bfseries,text=FFBlue!88!black] at ({.5*\scaleX*\xmax},23.05) {Independent panel A};
  \node[font=\scriptsize\bfseries,text=RCTeal!88!black] at ({\panelB+.5*\scaleX*\xmax},23.05) {Independent panel B};

  \foreach \base in {0,\panelB}{
    \draw[draw=FFGrayDark,line width=.72pt] (\base,.45)--(\base,22.55);
    \foreach \xx in {1,2,3,4,5}{
      \pgfmathsetmacro{\gx}{\base+\scaleX*\xx}
      \draw[FFRule] (\gx,.45)--(\gx,22.55);
    }
    \foreach \xx in {0,1,2,3,4,5}{
      \pgfmathsetmacro{\tx}{\base+\scaleX*\xx}
      \node[anchor=north] at (\tx,.18) {\xx};
    }
  }

  \node[anchor=east,font=\scriptsize\bfseries,text=FFGrayDark] at (-1.30,21.50) {Fixed actions};
  \node[anchor=east,font=\scriptsize\bfseries,text=FFGrayDark] at (-1.30,18.00) {Position};
  \node[anchor=east,font=\scriptsize\bfseries,text=FFGrayDark] at (-1.30,13.00) {Order};
  \node[anchor=east,font=\scriptsize\bfseries,text=FFGrayDark] at (-1.30,8.00) {Anchor};
  \node[anchor=east,font=\scriptsize\bfseries,text=FFGrayDark] at (-1.30,3.00) {H0};

  \foreach \yy/\lab/\lone/\mone/\ltwo/\mtwo in {
    22/{KEEP}/1.24161/2.25418/1.40889/2.40289,
    21/{XI}/3.45440/4.50932/4.12768/5.20516,
    20/{Ridge .1}/.67849/1.47910/.84712/1.68669,
    19/{Ridge 10}/.65910/1.45953/.78612/1.60832,
    18/{Ridge 1000}/.60229/1.43969/1.02286/1.89813,
    17/{Logit .1}/.53540/1.32405/.90612/1.73212,
    16/{Logit 10}/.44948/1.21368/.70720/1.51045,
    15/{Ridge .1}/.53755/1.32288/.76795/1.61016,
    14/{Ridge 10}/.54773/1.33243/.73087/1.57147,
    13/{Ridge 1000}/.45424/1.25156/.79808/1.65680,
    12/{Logit .1}/.74917/1.54948/.83077/1.63726,
    11/{Logit 10}/.62334/1.40604/.69209/1.48786,
    10/{Ridge .1}/.68102/1.46939/.73234/1.56460,
     9/{Ridge 10}/.55750/1.36566/.74246/1.57461,
     8/{Ridge 1000}/.86060/1.75050/.70408/1.56041,
     7/{Logit .1}/.60807/1.38416/.87476/1.69003,
     6/{Logit 10}/.68538/1.50808/.76023/1.58673,
     5/{Ridge .1}/.61545/1.42838/1.13729/2.04080,
     4/{Ridge 10}/.70403/1.52148/1.12347/2.01227,
     3/{Ridge 1000}/.64568/1.51298/.95604/1.85611,
     2/{Logit .1}/.74666/1.54693/1.00052/1.83288,
     1/{Logit 10}/.71463/1.54940/1.00192/1.83492
  }{
    \node[anchor=east] at (-.08,\yy) {\lab};

    \pgfmathsetmacro{\lA}{\scaleX*\lone}
    \pgfmathsetmacro{\mA}{\scaleX*\mone}
    \draw[draw=FFBlue,line width=.86pt] (\lA,\yy)--(\mA,\yy);
    \draw[draw=FFBlue,line width=.65pt,fill=white] (\lA-.055,\yy-.055) rectangle +(0.11,0.11);
    \fill[FFBlue] (\mA,\yy) circle[radius=.082];

    \pgfmathsetmacro{\lB}{\panelB+\scaleX*\ltwo}
    \pgfmathsetmacro{\mB}{\panelB+\scaleX*\mtwo}
    \draw[draw=RCTeal,line width=.86pt] (\lB,\yy)--(\mB,\yy);
    \draw[draw=RCTeal,line width=.65pt,fill=white] (\lB-.055,\yy-.055) rectangle +(0.11,0.11);
    \fill[RCTeal] (\mB,\yy) circle[radius=.082];
  }

  \foreach \yy in {20.5,15.5,10.5,5.5}{
    \draw[FFRule] (-2.55,\yy)--(8.72,\yy);
  }

  \node[font=\scriptsize] at (3.95,-1.42) {H8-policy minus current-policy utility ($\times10^{-4}$)};
\end{tikzpicture}}
\caption{Two independent Qwen2.5-7B panels across all 22 current-information comparators. Open squares mark simultaneous max-$t$ 95\% lower confidence bounds and filled circles mark realized point margins. Every lower-bound endpoint remains strictly positive; the two panels are not pooled.}
\label{fig:finite-library-replications}
\end{figure}
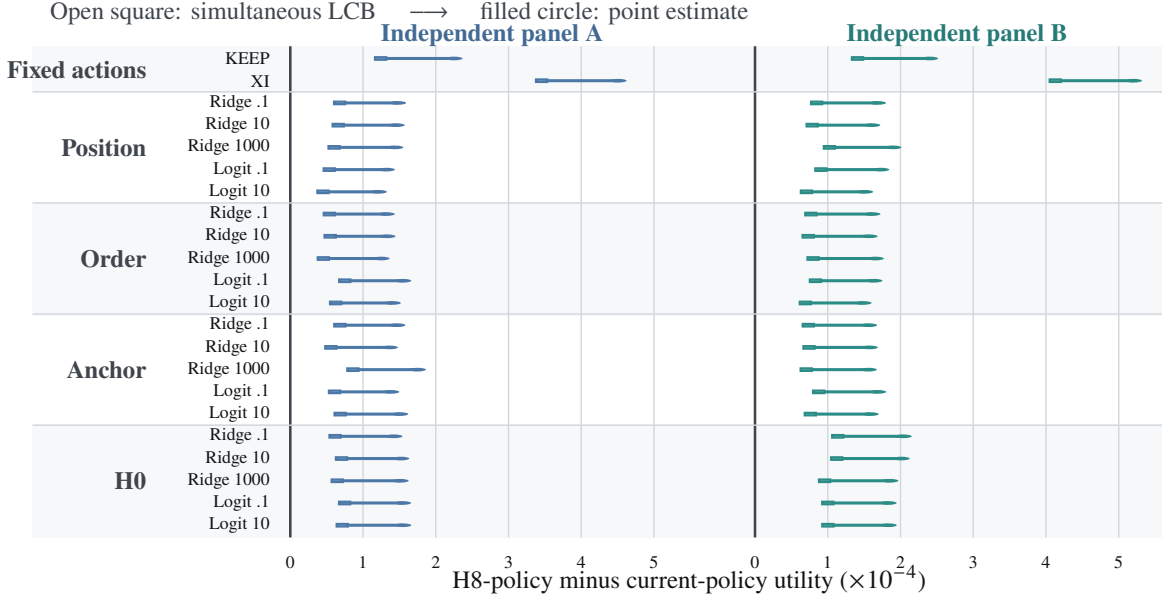

Before turning to the independent model-family replication, we first establish that the selected H8 future-response representation carries decision value beyond a broad declared current-information library in Qwen2.5-7B. The same prespecified 24-dimensional H8 raw-response Ridge policy ($\alpha=10$) is evaluated on two independent panels, each containing 336 exact anchors with exact position balance and an analysis fixed before target evaluation. The comparator library contains 22 legal current-information components: the two fixed actions plus five prespecified decision heads within each of four current-information feature families (position, order, anchor, and H0), as displayed in Figure~\ref{fig:finite-library-replications}. The two panels are analyzed independently rather than pooled.

\begin{table}[H]
\centering
\caption{Independent Qwen2.5-7B validation against the 22-component current-information library. Each entry is the worst component across the declared comparator set; positive lower bounds are required simultaneously.}
\label{tab:finite-library-replication}
\begin{tabular}{lcc}
\toprule
Quantity & Independent panel A (336) & Independent panel B (336)\\
\midrule
Minimum point margin & $1.2136812\!\times\!10^{-4}$ & $1.4878650\!\times\!10^{-4}$\\
Minimum Student-$t$ LCB & $6.9363323\!\times\!10^{-5}$ & $9.4261533\!\times\!10^{-5}$\\
Minimum bootstrap LCB & $7.0252125\!\times\!10^{-5}$ & $9.5137645\!\times\!10^{-5}$\\
Minimum max-$t$ LCB & $4.4947532\!\times\!10^{-5}$ & $6.9209070\!\times\!10^{-5}$\\
Comparators satisfying criterion & 22/22 & 22/22\\
\bottomrule
\end{tabular}
\end{table}

Every componentwise comparison remains positive in both panels for both point margins and simultaneous max-$t$ lower bounds; Figure~\ref{fig:finite-library-replications} shows all 22 comparisons on a common scale.

Productive-path value is also independently positive. Panel A gives
\begin{equation}
\widehat\Delta_{\rm path}^{(A)}=8.99131091346\times10^{-5},
\end{equation}
with one-sided Student-$t$ and bootstrap lower bounds $4.22948353683\times10^{-5}$ and $4.41065688619\times10^{-5}$. Panel B independently gives
\begin{equation}
\widehat\Delta_{\rm path}^{(B)}=1.47763681856726\times10^{-4},
\end{equation}
with corresponding lower bounds $1.01800375097654\times10^{-4}$ and $1.03042544792948\times10^{-4}$. These path values are later combined one panel at a time with the separate 432-anchor common-restart comparison through Equation~\eqref{eq:end-to-end-decomp}. The empirical claim in this section remains limited to the declared comparator library and the stated same-path execution technology.

\section{Frontier comparison: DynaMiCS-style probing}
\label{sec:frontier}

DynaMiCS performs short domain-specific fine-tuning probes to estimate a slope matrix of local cross-domain effects and then optimizes mixture weights under performance constraints \citep{gualdoni2026dynamics}. That makes it direct contemporary prior art for using future training itself as a prospective diagnostic. Our comparator is therefore not presented as the original DynaMiCS algorithm verbatim; it is a strengthened, task-adapted \emph{DynaMiCS-style} frontier that preserves the relevant short-probe/local-slope/restart geometry while using a calibrated binary decision head tailored to the present terminal action problem.

\paragraph{Why strengthen the comparator?}
The source algorithm solves a constrained mixture-selection problem, whereas our terminal decision is binary. A literal transplant therefore requires an additional mapping from local slope information to the present action gap. If that mapping were deliberately weak, a positive frontier result could reflect head misspecification rather than a limitation of short-probe information. We instead retain the source method's short-probe acquisition structure---same anchor, temporary short probes, local finite-difference summaries, restore/restart---but give the frontier a decision head fit only on development data for the declared H12 action gap. This strengthens the competing decision rule without granting deeper or otherwise illegal information.

For Qwen2.5-7B, the legal H4 feature vector is
\begin{equation}
z_F=
(L_A^0,L_B^0,L_C^0,
S_{\KEEP,A}^4,S_{\KEEP,B}^4,S_{\KEEP,C}^4,
S_{\XI,A}^4,S_{\XI,B}^4,S_{\XI,C}^4),
\end{equation}
where
\begin{equation}
S_{a,d}^4=
\frac{L_d(\theta_a^{H4})-L_d(\theta_0)}{4}.
\end{equation}
The Qwen decision head is fixed as \texttt{StandardScaler + Ridge($\alpha=0.1$)} with zero threshold and no test refit. Mistral-7B-v0.3 uses the same legal short-probe information class and restart technology with a model-specific numerical head fit on development data and fixed before target evaluation.

\subsection{Equalized compute contract}

Both methods receive exactly 20 policy-visible training updates:
\begin{center}
\begin{tabular}{lcc}
\toprule
Resource & Productive revelation & DynaMiCS-style restart\\
\midrule
Probe updates & 16 & 8\\
Fresh/promoted continuation & 4 & 12\\
Total policy-visible updates & 20 & 20\\
Discarded exploratory updates & 8 & 8\\
Reused selected-probe updates & 8 & 0\\
Terminal horizon & H12 & H12\\
Terminal utility & identical & identical\\
\bottomrule
\end{tabular}
\end{center}
Restore/save overhead is not charged against the comparator in the primary analysis, making the update accounting conservative for the productive-revelation method.

\paragraph{Theory-level comparison.}
Corollary~\ref{cor:depth-separation} gives a conditional separation from an H4 short-probe information class: if a decision-changing direction is invisible through H4 yet revealed by H8, then the deeper refinement can have strictly larger Bayes value at the same update budget because productive work is reusable. The empirical sections below test the corresponding structural refinement and end-to-end utility consequences in both model families.

\section{Structural decision refinement in the H4 frontier-ambiguity regime}
\label{sec:structural-refinement}

The 432-anchor comparative panel is disjoint from development and from the two finite-library replication panels.  Before terminal outcomes, the H4 frontier score defined the fixed ambiguity event
\begin{equation}
A=\{|f_F|\le\tau_F\},
\qquad
\tau_F=0.00052099142651661841.
\label{eq:ambiguity-final}
\end{equation}
The event contains 226/432 anchors, or 52.315\% of the panel.  Let
\begin{equation}
D_r=U_{T_r}(\XI)-U_{T_r}(\KEEP),\qquad r\in\{1,2\},
\end{equation}
and let $I_{\AR}$ be the fixed H8 active decision.  The two terminal banks are independent conditional replicates and expose both potential actions for every anchor.

\subsection{Independent-bank refinement estimator}

To test whether the H8 observation refines the coarse ambiguity cell in a decision-relevant way without reusing terminal outcomes for both choice and evaluation, bank $T_1$ chooses the best coarse constant action inside $A$,
\begin{equation}
\widehat c_1=\ind\{\overline D_{1,A}>0\},
\end{equation}
and the independent bank $T_2$ evaluates the fixed H8 decisions relative to that coarse action:
\begin{equation}
\widehat\Delta_{1\rightarrow2}(A)
=
\frac{1}{|A|}\sum_{i\in A}D_{2i}(I_{\AR,i}-\widehat c_1).
\label{eq:cross-refine-12}
\end{equation}
The reverse direction $\widehat\Delta_{2\rightarrow1}(A)$ is defined symmetrically, and the primary symmetric estimator is their average.  In each exact-anchor bootstrap draw, the coarse action is reselected inside the resample before evaluation.

The two directions agree closely:
\begin{align}
\widehat\Delta_{1\rightarrow2}(A)&=1.92863119513\times10^{-4},
&\mathrm{LCB}_{t}&=5.01319646865\times10^{-5},\\
\widehat\Delta_{2\rightarrow1}(A)&=1.93079378050\times10^{-4},
&\mathrm{LCB}_{t}&=1.03136801205\times10^{-4}.
\end{align}
Their symmetric average is
\begin{equation}
\boxed{
\widehat\Delta_{\rm refine}(A)=1.92971248782\times10^{-4},
}
\end{equation}
with one-sided 95\% Student-$t$ lower bound $1.09966903938\times10^{-4}$ and 50,000-draw exact-anchor bootstrap lower bound $6.49311403490\times10^{-5}$.  Weighting by the fixed ambiguity mass gives a full-panel contribution $1.00952551446\times10^{-4}$, with $t$ lower bound $5.69892792476\times10^{-5}$ and bootstrap lower bound $3.39715294723\times10^{-5}$.

\subsection{Opposite-action sign replication}

The mechanism is visible independently in both terminal banks.  Table~\ref{tab:sign-replication} reports the terminal action gap conditional on the fixed H8 action within $A$.

\begin{table}[H]
\centering
\small
\caption{Independent-bank structural sign replication inside the fixed H4 frontier-ambiguity event.  Bounds are one-sided 95\% Student-$t$ bounds.}
\label{tab:sign-replication}
\begin{tabular}{lrrrr}
\toprule
Bank & $\XI$-selected mean & LCB & $\KEEP$-selected mean & UCB\\
\midrule
$T_1$ & $2.9246\times10^{-4}$ & $6.5097\times10^{-5}$ & $-4.1166\times10^{-4}$ & $-2.2471\times10^{-4}$\\
$T_2$ & $3.6323\times10^{-4}$ & $9.5547\times10^{-5}$ & $-2.8389\times10^{-4}$ & $-8.2269\times10^{-5}$\\
\bottomrule
\end{tabular}
\end{table}

Thus the H8 observation does not merely predict a continuous outcome more accurately: within a materially populated H4 frontier-ambiguity regime it reproducibly separates groups of states whose mean independent terminal outcomes favor opposite actions.  Treating the ambiguity event as the coarse cell and the fixed H8 partition as the refinement gives a sample Bayes gap of $1.63115121441\times10^{-4}$ using the mean of $T_1,T_2$, with bootstrap lower bound $6.04584733171\times10^{-5}$.  The same cell-refinement Bayes gap has a positive bootstrap lower bound in each terminal bank separately.

\paragraph{Secondary H4-versus-H8 diagnostic.}
A cross-bank diagnostic that preserves outcome separation compares the fixed nine-feature H4 Ridge representation with the fixed 24-feature H8 representation for terminal-gap prediction.  Inside $A$, the H8 representation improves cross-bank mean squared error by $1.24677\times10^{-7}$, with positive $t$ and bootstrap lower bounds.  The corresponding learned-policy value difference has a confidence interval crossing zero, reflecting that the strengthened H4 frontier action is already a strong selector.  We therefore interpret the main result as \emph{decision-relevant H8 refinement inside the declared H4 frontier-ambiguity regime}, not as unrestricted dominance over every measurable function of complete H4 information.

\subsection{Independent Mistral aliasing and deeper-revelation value}
\label{sec:mistral-structural}

The independent Mistral target panel contains 480 exact anchors with two terminal banks. Its H4 ambiguity threshold, $\tau_{F}^{\mathrm M}=0.00236920914414$, was determined from the separate development panel. Using Bank A's H4 score to define the ambiguity cell yields 216/480 anchors; among them, 88 favor $\XI$ in both terminal banks and 58 favor $\KEEP$ in both. Defining the cell with Bank B gives 210/480 anchors, with 83 stable-$\XI$ and 58 stable-$\KEEP$ anchors. Thus the same coarse shallow-information regime contains materially populated hidden-state subsets requiring opposite terminal actions.

More directly, the fixed H8 action has strictly positive terminal decision value relative to the fixed H4 action on the complete two-bank target panel. The exact-anchor mean is
\begin{equation}
\boxed{\widehat\Delta_{H8-H4}^{\mathrm M}=2.25089539402\times10^{-4},}
\end{equation}
with one-sided 95\% Student-$t$ lower bound $1.56391515827\times10^{-4}$ and 50,000-draw bootstrap lower bound $1.58226436880\times10^{-4}$. H4 and H8 actions differ on 100/480 exact anchors in at least one bank. Mistral therefore independently reproduces the core implication of decision-revelation depth: legal deeper future-learning information changes action choice on nontrivial mass and has strictly positive terminal decision value.

A stronger Qwen diagnostic in which the H8-selected partition itself must produce opposite-sign subgroup bounds in both directions is not promoted as a cross-model invariant. The theorem requires decision-relevant refinement, not numerical identity of every diagnostic partition.

\section{Productive revelation and equal-compute frontier performance}
\label{sec:frontier-performance}

We next ask whether additional revelation can be converted into higher end-to-end utility than the strengthened DynaMiCS-style H4 restart frontier under the common 20-update contract. Equation~\eqref{eq:end-to-end-decomp} gives the exact decomposition
\begin{equation}
\Delta_{\rm end}
=
\Delta_{\rm restart}^{\AR-\Frontier}
+
\Delta_{\rm path}^{\AR}.
\label{eq:frontier-component-decomp}
\end{equation}
This separates what is gained by choosing better from what is gained because the selected deeper probe remains useful computation.

\subsection{Qwen equal-compute closure}

The 432-anchor Qwen panel estimates the common-restart selector component using the mean of two independent terminal banks:
\begin{equation}
\widehat\Delta_{\rm restart}^{\AR-\Frontier}
=-3.98358746688\times10^{-7},
\end{equation}
with standard deviation $4.67246\times10^{-4}$. The point estimate is therefore near zero; this is descriptive rather than an equivalence claim. Strict end-to-end advantage is supplied by productive reuse of the tested H8 path.

Combining the independent common-restart estimate separately with the two productive-path panels gives
\begin{align}
\widehat\Delta_{\rm end}^{(1)}&=8.95147503879\times10^{-5},
&\mathrm{LCB}_{0.95}&=2.92463884692\times10^{-5}>0,\\
\widehat\Delta_{\rm end}^{(2)}&=1.47365323110\times10^{-4},
&\mathrm{LCB}_{0.95}&=8.83939050151\times10^{-5}>0.
\end{align}
The productive-path values themselves are $8.9913\times10^{-5}$ and $1.4776\times10^{-4}$, respectively, with positive one-sided $t$ and bootstrap lower bounds in both panels.

\subsection{Independent Mistral equal-compute replication}

The same decomposition is evaluated on the independent Mistral target panel under the identical 20-update visible-compute contract. The common-restart selector component is
\begin{equation}
\widehat\Delta_{\rm restart}^{\mathrm M}=-1.191350097\times10^{-5},
\end{equation}
whereas same-path reuse contributes
\begin{equation}
\boxed{\widehat\Delta_{\rm path}^{\mathrm M}=3.09392744795\times10^{-4},}
\end{equation}
with one-sided Student-$t$ and bootstrap lower bounds $1.96017743591\times10^{-4}$ and $2.01665987607\times10^{-4}$. The resulting end-to-end equal-compute advantage is
\begin{equation}
\boxed{\widehat\Delta_{\rm end}^{\mathrm M}=2.97479243825\times10^{-4},}
\end{equation}
with one-sided Student-$t$ lower bound $2.00162462592\times10^{-4}$ and bootstrap lower bound $2.04870462705\times10^{-4}$.

\begin{table}[H]
\centering
\caption{Equal-compute productive-revelation evidence across the two Transformer families. The two Qwen rows use independent productive-path panels and are not pooled.}
\label{tab:frontier-crossmodel}
\small
\begin{tabular}{lrrr}
\toprule
Model / evidence block & Productive path & End-to-end point & One-sided 95\% LCB\\
\midrule
Qwen, independent panel 1 & $8.9913\times10^{-5}$ & $8.9515\times10^{-5}$ & $2.9246\times10^{-5}$\\
Qwen, independent panel 2 & $1.4776\times10^{-4}$ & $1.4737\times10^{-4}$ & $8.8394\times10^{-5}$\\
Mistral, independent family & $3.0939\times10^{-4}$ & $2.9748\times10^{-4}$ & $2.0016\times10^{-4}$\\
\bottomrule
\end{tabular}
\end{table}

The same decomposition is supported in both families: the restart-only selector difference is not the source of the main gain, while productive reuse of deeper tested computation creates a strict equal-budget end-to-end advantage. This cross-family pattern is consistent with the productive-revelation mechanism formalized in Equation~\eqref{eq:end-to-end-decomp}.

\section{Adaptive Revelation Control across model families}
\label{sec:adaptive-validation}

The dynamic theory predicts a state-dependent stopping decision based on the conditional utility of deeper revelation, not a universal population ambiguity threshold. The empirical question is therefore whether legal shallow information can support compute-saving continuation decisions without changing the terminal utility definition or productive compute ledger. The theory also permits different model families to occupy different conditional revealability regimes.

\subsection{Qwen2.5-7B: decision-nonredundant revealability and held-out safe compute}
\label{sec:adapt2-two-bank}

In Qwen2.5-7B, the observable H4 revealability proxy is
\begin{equation}
R_4=
\left\|
\bigl(
S_{\XI,A}^4-S_{\KEEP,A}^4,
S_{\XI,B}^4-S_{\KEEP,B}^4,
S_{\XI,C}^4-S_{\KEEP,C}^4
\bigr)
\right\|_2.
\label{eq:qwen-r4}
\end{equation}
We treat $R_4$ only as a legal model-specific proxy; no equality $R_4=\sigma_4$ is assumed. In a supporting analysis with fitting and evaluation performed on separate collection blocks under the same safety protocol, the two-coordinate continuation rule improves paired cost-aware value over the scalar $|f_{H4}|$ rule by $3.2514\times10^{-7}$, with one-sided Student-$t$ and bootstrap lower bounds $4.0357\times10^{-8}$ and $3.6127\times10^{-8}$, and no harmful stops in either held-out direction. This is evidence for a decision-nonredundant regime at the declared compute price, not a prospective claim that this particular $R_4$ is unique or universally necessary.

A separate 480-history full-grid study on new exact learning-state anchors prospectively fixed the utility-aware continuation architecture, H4/H8 policies, update price, and 16/20-update compute ledger. With an independent second future bank on the same anchors, the H8-over-H4 terminal decision value is
\begin{equation}
\widehat{\E}[Q_{\pi_8}-Q_{\pi_4}]_{AB}=1.229\times10^{-4},
\qquad
\mathrm{LCB}_{t,0.95}=8.687\times10^{-5}>0,
\qquad
\mathrm{LCB}_{\mathrm{boot},0.95}=8.769\times10^{-5}>0.
\label{eq:adapt2-deep-value}
\end{equation}
The fixed two-coordinate point controller has positive two-bank net point value but its strict one-sided net-value lower bounds cross zero, illustrating the finite-sample tail sensitivity emphasized by the theory.

The bounded-risk certification layer provides a complementary held-out safe-compute result. In a theory-constrained calibration/held-out evaluation, one bank is used for calibration and the other for evaluation. The resulting calibrated rule stops on 4/480 histories, produces no H4/H8 action changes and no harmful stops, has a 95\% Bernoulli--KL upper joint-risk bound of $0.6222\%<2\%$, and saves $0.1667\%$ of visible updates. Its realized panel-level update-priced net value is
\begin{equation}
\widehat{\Delta V}_{\rm cert}=2.890\times10^{-7},
\qquad
\mathrm{LCB}_{t,0.95}=5.161\times10^{-8}>0,
\qquad
\mathrm{LCB}_{\mathrm{boot},0.95}=7.225\times10^{-8}>0.
\label{eq:adapt2-certified-net}
\end{equation}

\subsection{Mistral-7B-v0.3: scalar continuation control at the tested resolution}
\label{sec:mistral-adaptive}

Mistral-7B-v0.3 does not require numerical transfer of Qwen's $R_4$. A Mistral-specific continuation head is instantiated using only the separate 336-anchor development panel, while the independent 480-anchor two-bank panel is used for target evaluation without target-panel refitting. Among the two pre-existing shallow continuation architectures evaluated as a finite family, the scalar margin-based estimator retains positive adaptive value at the observed resolution.

Relative to always continuing to H8, the scalar utility-aware controller achieves exact-anchor update-priced net value
\begin{equation}
\boxed{\widehat{\Delta V}_{\rm scalar}^{\mathrm M}=1.13877630257\times10^{-5}.}
\end{equation}
Its ordinary one-sided 95\% Student-$t$ and 50,000-draw bootstrap lower bounds are $2.37141363069\times10^{-6}$ and $5.41900958243\times10^{-6}$. Because both scalar and two-coordinate continuation architectures existed before the Mistral target evaluation, we treat them as a two-element finite architecture family. Bonferroni-adjusted familywise 95\% lower bounds for the scalar rule remain positive: $6.37735915928\times10^{-7}$ for Student-$t$ inference and $5.23837592968\times10^{-6}$ for the exact-anchor bootstrap. Across 960 bank-level realizations it stops 166 times, changes the H8 action once, and has zero harmful stops under the declared update price. The raw Qwen-style $R_4$ coordinate does not improve Mistral adaptive value over the scalar controller; exploratory development-only proxy diagnostics are excluded from the target-panel claim. We therefore interpret Mistral as consistent with scalar decision sufficiency under Theorem~\ref{thm:scalar-control-factorization} and Corollary~\ref{cor:cost-adjusted-revealability-crossing}, rather than as a failure of the two-coordinate theorem. This empirical conclusion does not assert that $\sigma_4$ is literally a function of margin; it says only that the finite pre-existing architecture family did not require a second coordinate to improve the priced target-panel stopping decision at the tested resolution.

\subsection{Risk--severity separation in Mistral}

A separate risk-calibrated Mistral shallow-stop rule provides an empirical illustration of Theorem~\ref{thm:risk-only-impossibility}. It makes 351/960 early stops while the 95\% Bernoulli--KL joint stop/action-change upper bound is $1.5924\%<2\%$ in each bank. Yet the exact-anchor update-priced net point value is slightly negative,
\begin{equation}
\widehat{\Delta V}_{\rm risk}^{\mathrm M}=-3.3428\times10^{-7},
\end{equation}
with negative one-sided lower bounds. Only four stopped action changes occur, but the largest absolute continuation value is $9.811\times10^{-3}$. Thus low action-instability probability and positive utility are empirically distinct objects: rare high-severity misses can dominate many cheap correct stops. This is precisely the risk--severity separation formalized in Appendix~\ref{app:adaptive-cert}.

Taken together, the two model families support the same adaptive-control principle while occupying different shallow-state regimes. Qwen exhibits evidence for a decision-nonredundant revealability coordinate; Mistral admits a successful scalar continuation rule. What transports is the conditional revelation-value decision, not a universal empirical coordinate system.

\section{Cross-model structural validation}
\label{sec:crossmodel-synthesis}

The two model families are not expected to reproduce identical coefficients, thresholds, or empirical revealability coordinates. The relevant replication object is the sequence of theory-level implications: shallow information can alias decision-relevant state; deeper revelation can have positive decision value; productive reuse can convert that information into equal-compute utility; revelation depth can be controlled state by state; and risk control is not equivalent to utility control when severity is unrestricted.

\begin{table}[H]
\centering
\caption{Theory-level evidence across the two Transformer families. The fixed-depth and equal-compute Mistral results use an independent target panel; the adaptive row evaluates a pre-existing two-architecture family with familywise-adjusted inference.}
\label{tab:crossmodel-synthesis}
\small
\begin{tabularx}{\linewidth}{L{0.25\linewidth}YY}
\toprule
Theoretical implication & Qwen2.5-7B evidence & Mistral-7B-v0.3 evidence\\
\midrule
Decision-relevant shallow aliasing
& Fixed H4 ambiguity cell contains H8-refined groups with opposite terminal action gaps in two independent banks.
& Using the H4 threshold fixed from development data, the bank-A-defined ambiguity cell contains 216/480 anchors, including 88 stable-$\XI$ and 58 stable-$\KEEP$ anchors whose terminal-gap signs agree across both banks (descriptive structural check).\\
\addlinespace
Positive value of deeper revelation
& H8 beats the declared current-information library on two independent 336-anchor panels; the separate 480-history H8-over-H4 value is strictly positive.
& Exact-anchor H8-over-H4 value $2.2509\times10^{-4}$ with positive $t$ and bootstrap lower bounds.\\
\addlinespace
Productive information-to-utility closure
& Productive path value is positive in two independent panels and yields strict 20-update end-to-end advantage over the strengthened restart frontier.
& Same-path reuse $3.0939\times10^{-4}$ and equal-compute end-to-end advantage $2.9748\times10^{-4}$, both with positive lower bounds.\\
\addlinespace
Adaptive Revelation Control
& Supporting evidence for incremental decision value from a nonredundant revealability coordinate; a held-out bounded-risk safe-compute evaluation has positive panel net-value lower bounds.
& Consistent with scalar decision sufficiency at the tested resolution; a familywise structural validation of the scalar continuation architecture, fit only on the separate development panel, retains positive Bonferroni-adjusted exact-anchor $t$ and bootstrap lower bounds within the two-element pre-existing architecture family.\\
\addlinespace
Risk--severity boundary
& Positive held-out safe-compute behavior is explicitly limited to panel-level evidence without an unrestricted tail guarantee.
& Sub-2\% stop--flip risk coexists with nonpositive risk-controller net value because a few flips have high severity.\\
\bottomrule
\end{tabularx}
\end{table}

The common pattern is therefore stronger than numerical parameter reuse. In both families, future learning supplies decision-relevant information, deeper tested paths carry productive value, and the resulting information can improve compute-accounted utility. The model-specific difference occurs one level lower: Qwen provides evidence of incremental decision value from an additional shallow revealability coordinate, whereas Mistral is consistent with scalar decision sufficiency at the observed resolution. Theorem~\ref{thm:scalar-control-factorization} shows why these outcomes belong to the same theory: an extra coordinate is required for control only when states sharing the scalar summary cross the priced \textsc{Stop}/\textsc{Continue} boundary. Corollary~\ref{cor:cost-adjusted-revealability-crossing} gives the corresponding location--scale criterion, while Definition~\ref{def:model-specific-revealability-proxy} allows model-specific empirical realizations.

Accordingly, the cross-model conclusion is not that one fitted controller or proxy transfers unchanged. It is that the \emph{Revelation-Control relation} between legal shallow information, continuation value, productive cost, and terminal decision utility survives an independent model-family change. This is the level at which the empirical evidence supports the general theory's central testable implications.

\section{Potential application domains}
\label{sec:applications}

The empirical results in this paper concern learning systems, but the decision-theoretic structure is more general. A Revelation-Control problem can be formulated when five ingredients can be defined: a consequential terminal action; hidden state not fully resolved by current legal information; a priced intervention that can be executed before commitment; an observable response that can refine the relevant decision quotient; and a cost ledger that distinguishes information acquisition from any useful state change left by the intervention. The strongest form of the framework arises when the experiment is \emph{productive}: if the tested path is selected, some of the work performed to reveal state is itself part of the useful continuation. These conditions are substantially narrower than ``any sequential decision problem,'' and the examples below are prospective application domains rather than deployment claims established by the present experiments.

\subsection{Learning and computational systems}

\paragraph{Foundation-model adaptation and data-domain allocation.}
Fine-tuning a large model often requires choosing among domains, task mixtures, adapters, or continuation depths under a limited compute budget. Two checkpoints can look similar under current loss or held-out behavior yet respond differently to the same candidate domain update. A short domain-specific training intervention can therefore serve as a future-learning experiment: its response can refine the decision quotient over which domain or adapter should be promoted, while the selected probe trajectory can remain useful training. This differs from using a learning curve only to forecast final loss; the target is whether newly revealed learning state changes a downstream adaptation decision. The DynaMiCS-style comparison studied here is one concrete instance of this broader allocation problem \citep{gualdoni2026dynamics}.

\paragraph{Continual learning, curriculum control, rehearsal allocation, and AutoML.}
In continual or curriculum learning, present task performance need not reveal how much plasticity, interference, or recoverability remains for the next phase. A bounded amount of rehearsal or target-task training can be treated as an experiment on that hidden state before the curriculum decision is fixed. The same principle applies to hyperparameter and configuration search: partial-training methods already pause, terminate, or promote candidate runs from intermediate performance \citep{swersky2014freeze,domhan2015speeding,li2018hyperband,jaderberg2017pbt}, but Revelation Control asks whether additional computation should be purchased because its \emph{response} is expected to change the decision. A run with a mediocre current metric can justify continuation when deeper revelation has positive decision value, while a promising-looking run can be stopped when additional information is unlikely to alter the selected action. For long-horizon or lifelong learning, an important extension is to identify a minimal revelation state that remains dynamically sufficient across successive stages; if such a state satisfies multi-stage Markov closure, Revelation Control could be composed recursively rather than applied only as a one-step continuation rule. Productive promotion further distinguishes useful exploratory work from discarded evaluation cost.

\paragraph{Intervention-aware data acquisition.}
The intervention can also be a training batch, synthetic-data block, augmentation regime, rehearsal set, or other legal update whose response is observable before a larger commitment is made. Classical Bayesian experimental design and active sensing choose observations by expected decision value \citep{lindley1956measure,chaloner1995bayesian,veiga2023active}. In a learning system, the experiment can instead be an update to the learner itself. Development-only future responses can be used to estimate a model-specific revealability coordinate or continuation state and to determine whether current information is already decision-sufficient. This suggests a route to choosing not merely which data are informative, but which data reveal something about the learner's \emph{future response} that changes the action menu.

\paragraph{Adaptive computation beyond training.}
The depth variable need not literally count gradient updates. In an iterative solver, multi-fidelity simulation, branch-and-bound routine, Monte Carlo computation, or other staged numerical procedure, a coarse computation may be insufficient to choose a downstream action while additional computation both reveals information and accumulates reusable work. If the current approximation, the refinement response, and the terminal decision can be placed in a common utility ledger, revelation depth becomes computation depth. The same continuation principle then asks whether another refinement level is worth purchasing because of the decision it may change, connecting Revelation Control to value-of-computation ideas \citep{russell1991principles,frazier2008knowledge} while retaining the productive-reuse term.

\subsection{Beyond machine learning: productive experiments in sequential decisions}

\paragraph{Control, system identification, and robotics.}
A physical system can occupy internal states with nearly identical current outputs but different future responses and therefore different optimal controls. Examples include uncertain friction, payload, contact mode, actuator health, or local dynamics. Dual control already formalizes the fact that an action can simultaneously control a system and reveal uncertainty \citep{feldbaum1960dual}, while POMDP and active-sensing formulations treat task-directed information acquisition under partial observability \citep{kaelbling1998planning,veiga2023active}. Revelation Control suggests a decision-relative refinement of this idea when a short controlled motion or excitation can be priced, its response can reveal which control action is appropriate, and the selected probing trajectory can be retained rather than discarded. The theory does not require complete system identification: only the hidden distinctions that change the terminal action need to be resolved.

\paragraph{Industrial process control, diagnostics, and maintenance.}
Machines or production processes can present similar current telemetry while differing in latent wear, loading, material state, or failure proximity. A short controlled operating segment, calibration cycle, or diagnostic excitation can reveal whether the correct action is to continue production, derate, switch operating mode, inspect, or maintain. Such settings are especially close to productive revelation when the diagnostic segment also performs useful work. The relevant accounting must, however, include downtime, safety margin, irreversible wear, and state-restoration cost; if probing itself damages the system or changes the target regime, the promotion geometry used in the present experiments no longer applies without modification.

\paragraph{Scientific experimentation and automated laboratories.}
In materials, chemistry, biology, and other experimental sciences, an early observation may leave several latent mechanisms compatible with the data even though those mechanisms imply different next experiments or operating decisions. Bayesian experimental design already provides a mature decision-theoretic language for choosing informative experiments \citep{lindley1956measure,chaloner1995bayesian}. Revelation Control is potentially relevant when an experiment can be staged: an initial intervention is run only far enough to determine whether deeper measurement or continuation is decision-relevant, and the partial trajectory can be retained if that branch is selected. Automated laboratories, adaptive measurement pipelines, and multi-stage physical experiments are natural settings in which ``how much experiment to purchase'' may be as important as ``which experiment to run.'' Establishing this connection rigorously would require domain-specific transition models, measurement error, and experimental-cost accounting beyond the present paper.

\paragraph{Operations, resource allocation, and pilot-to-scale decisions.}
Many operational decisions involve latent regimes that cannot be resolved from passive current data alone. A limited allocation, pilot deployment, test market, routing perturbation, or operating-policy trial can reveal response heterogeneity before a larger commitment is made. The Revelation-Control abstraction is appropriate when the pilot has a well-defined terminal decision, its response can be evaluated before scale-up, and at least part of the pilot's value or infrastructure is reusable if promoted. The central distinction is between \emph{learning about the world} and \emph{learning enough to choose the action}: a pilot need not identify the full latent regime if it already resolves the relevant decision quotient. Strategic behavior, interference, nonstationarity, and general-equilibrium effects can break the simple controlled-probe assumptions and would have to be modeled explicitly.

\paragraph{High-stakes human decision systems.}
The mathematical pattern also appears, at least abstractly, in adaptive treatment, public-policy pilots, and personalized education: two individuals or populations with similar current observables can respond differently to a small intervention, and the response can change the preferred next action. These domains are intentionally placed at the outer boundary of the present paper's application claims. A valid extension would require causal identification, ethical constraints, uncertainty about intervention harm, fairness considerations, and stronger safety guarantees than the bounded-risk analysis developed here. In medicine or policy especially, ``productive probing'' cannot be assumed merely because an intervention also has intended benefit; the value ledger must price adverse effects and irreversibility, and experimentation may be impermissible even when its statistical information value is high. The present work therefore supplies only a possible decision-theoretic template, not a deployment prescription.

\subsection{A domain-level applicability test}

The most useful transfer question is not whether a new field resembles model fine-tuning superficially, but whether its decision structure matches the theory. For a new system, one should first declare the legal current information, the terminal action menu, terminal utility, admissible probe interventions, and all costs or safety penalties. Development-only probe responses can then be used to estimate the conditional continuation value and to test whether the current scalar summary is already decision-sufficient. If scalar fibers cross the priced \textsc{Stop}/\textsc{Continue} boundary, additional revealability information is decision-nonredundant and a richer controller is required; if they do not, a simpler controller can be sufficient even when hidden-state variation remains. The resulting architecture and cost ledger should be fixed before target evaluation. If the probe leaves reusable work, information value and productive reuse must be reported separately; if it does not, the problem reduces to discard-and-restart information acquisition. This procedure is the general method that transfers across domains: the decision factorization, continuation-value criterion, and accounting principles are invariant, whereas observable proxies, dynamics, coefficients, safety constraints, and prices are system-specific.

\section{Limitations}

\paragraph{Scope of the empirical systems.}
The empirical study now spans two independently instantiated 7B decoder-only Transformer families, but both use LoRA adaptation, AdamW-style optimizer state, the same binary intervention menu, and the same H4/H8/H12 control geometry. Cross-optimizer, substantially different-scale, non-Transformer, and materially different-horizon generality remain open.

\paragraph{Comparator scope.}
The finite-library result does not imply dominance over every measurable current-information policy. The frontier comparison is against one strengthened DynaMiCS-style short-probe/restart class under a common policy-visible update budget, not against every possible probing or metareasoning method.

\paragraph{Productive revelation assumptions.}
Promotion is useful only when tested work remains legally and scientifically reusable. If a probe corrupts the deployment path, creates irreversible risk, changes the target distribution, or incurs additional state-I/O or safety cost, restart may be the appropriate technology and the budget identity must be repriced.

\paragraph{Conditional rather than universal revealability dimension.}
The location--scale theorem is a structural statement about $(|m_h|,\sigma_h)$, not a claim that every model must exhibit two empirically nonredundant shallow coordinates. The exact scalar-control factorization theorem further shows that residual variation in $\sigma_h$ need not be decision-relevant: a second coordinate is required only when a positive-mass scalar fiber crosses the priced \textsc{Stop}/\textsc{Continue} boundary. Qwen provides evidence for a decision-nonredundant regime, while Mistral is consistent with scalar decision sufficiency at the observed resolution. The empirical $R_4$ used in Qwen is one legal model-specific proxy, not a universal statistic.

\paragraph{Tail-robust utility certification.}
Controlling the probability that an early stop changes the deeper action does not by itself certify positive expected utility when stopped-flip severity is unrestricted. Theorem~\ref{thm:risk-only-impossibility} shows that this is an identification boundary rather than merely a power limitation: a population net-utility certificate additionally requires a predeclared severity, moment, or integrable-tail class. Positive held-out panel-level utility is therefore not a universal tail-robust guarantee.

\paragraph{Applications beyond learning systems.}
The mathematical decision structure can be instantiated whenever a priced controlled intervention reveals decision-relevant hidden state before commitment, but the empirical evidence in this paper is confined to learning systems. Physical, scientific, operational, medical, policy, and educational settings introduce domain-specific causality, irreversibility, strategic response, measurement error, safety, fairness, and ethical constraints that are not resolved here. The broader domains in \cref{sec:applications} are therefore prospective uses of the framework, not validated deployments or claims of domain-universal optimality.

\paragraph{Method optimality.}
The evaluated policies are selected using development data and fixed before target evaluation. No result establishes global optimality over all representations, learners, stopping rules, probe schedules, intervention menus, or acquisition policies.

\paragraph{Compute and deployment value.}
The main normalization counts policy-visible training updates under a common model and batch construction. Restore/save overhead is not charged against the frontier in the main comparison. Device-seconds, wall-clock, memory, state-I/O, monetary value, safety value, and live-deployment utility are different resources and are not collapsed into the primary claim.

\paragraph{Dynamic-state compression.}
The location--scale result is an exact one-step statement. One-step low-dimensional decision sufficiency does not imply multi-stage Markov closure. A multi-depth controller can therefore use $(m_h,\sigma_h)$ as a complete low-dimensional state only under an additional Markov-closure condition for future margin--revealability dynamics. Characterizing minimal dynamically sufficient revelation states, and the conditions under which they induce Markov closure across revelation depths, is a natural next theoretical problem.

\section{Conclusion}

The companion work established that present behavior can hide distinctions in future learning. The present paper closes the decision-theoretic step: such distinctions can have action value, can be actively revealed by controlled future learning, and can become a state-dependent control variable. We call this problem Revelation Control.

The theory identifies the relevant object as a decision quotient rather than the full hidden state. A coarse observation is sufficient only when the refined optimal action factorizes through it. The fork--probe--promote realization studied here uses future learning as an experiment on hidden learner state, while productive revelation keeps the selected tested path as useful computation. The resulting value ledger separates pure information gain from the value of already executed work. Static Bayes refinement value then closes exactly into a conditional revelation value, yielding an optimal-stopping rule over revelation depth.

The empirical evidence supports this structure in two independently instantiated Transformer families. In Qwen2.5-7B, a fixed H8 policy strictly beats a declared 22-component current-information library on two independent panels, refines a fixed H4 ambiguity regime into groups with opposite terminal action value, and obtains strict equal-compute advantage over a strengthened DynaMiCS-style H4 restart frontier through productive reuse. In Mistral-7B-v0.3, an independent two-bank panel reproduces positive H8-over-H4 decision value and the same productive equal-compute mechanism with positive one-sided lower bounds. The cross-model result is therefore not numerical reuse of one fitted policy: it is replication of the information-to-utility structure predicted by the theory.

Adaptive control exhibits the regime dependence predicted by the theory. Qwen provides evidence that a second shallow revealability coordinate can matter beyond current margin; a familywise-adjusted Mistral structural validation, restricted to the finite pre-existing architecture family and fit only on independent development data, supports a successful scalar continuation rule at the tested resolution. The exact scalar-control factorization result makes the common structure precise: an additional coordinate is necessary for the stopping decision only when states sharing the scalar summary fall on both sides of the cost-adjusted continuation boundary. Thus predictive or value variation can remain present without being decision-nonredundant. A separate Mistral risk-calibrated controller further illustrates the impossibility boundary: sub-2\% decision-instability risk can coexist with nonpositive net utility when a few stopped flips have large severity.

The resulting conclusion is specific but broad in implication. Hidden learning state is not merely predictive structure. Controlled future learning can reveal distinctions that change decisions; productive reuse can convert revelation into higher utility at the same visible training-update budget; and the amount of revelation can itself be controlled according to state-dependent expected value. More generally, \cref{sec:applications} identifies the same decision pattern in control and system identification, robotics and industrial diagnostics, scientific experimentation, adaptive computation, and pilot-to-scale operational decisions, with high-stakes human domains requiring additional causal, ethical, and safety theory. What must transport across systems is the decision structure and value accounting, not a universal proxy, coefficient, threshold, or even a fixed empirical state dimension. Revelation Control therefore provides a theory-first framework for deciding not only \emph{what} action to take, but \emph{how much controlled future interaction it is worth purchasing in order to know}.

\appendix
\section{Conservative replicate-denoised Bayes-regret bridge}

This appendix records an optional stronger certification device.  It is not required for the main empirical conclusions in Sections~\ref{sec:structural-refinement}--\ref{sec:frontier-performance}.  Independently generated future outcomes can be used to separate persistent conditional signal from one-bank realization noise and to upper-bound coarse-policy Bayes regret under the stated assumptions.

Consider binary actions with terminal gap
\begin{equation}
D=Q_{\XI}-Q_{\KEEP}.
\end{equation}
Let future banks $A$ and $B$ share a pre-bank sigma-field $\mathscr X$ and satisfy
\begin{equation}
\E[D_A\mid\mathscr X]
=
\E[D_B\mid\mathscr X]
=:\mu.
\end{equation}
The legal coarse information obeys $\HB\subseteq\mathscr X$, and
\begin{equation}
m_{\HB}=\E[\mu\mid\HB]
\end{equation}
provides the coarse-information Bayes score.  Let $f$ be any fixed $\HB$-measurable score in the same utility units as $D$, and let
\begin{equation}
\pi_f=\ind\{f>0\}.
\end{equation}
Here and below, action indicator one denotes $\XI$ and zero denotes $\KEEP$.

Define the cross-replicate residual moment
\begin{equation}
M_\times(f)=\E[(D_A-f)(D_B-f)].
\end{equation}
Let
\begin{equation}
c_{\mathscr X}
=
\operatorname{Cov}(D_A,D_B\mid\mathscr X).
\end{equation}
Conditional independence gives $c_{\mathscr X}=0$; allowing $c_{\mathscr X}\ge0$ yields the same conservative direction.

\begin{theorem}[Replicate-denoised Bayes-regret bridge]
Assume square integrability, a shared conditional mean across banks, and $c_{\mathscr X}\ge0$ almost surely.  Then
\begin{align}
M_\times(f)
&=
\E[(\mu-f)^2]+\E[c_{\mathscr X}] \\
&=
\E[(m_{\HB}-f)^2]
+
\E[\operatorname{Var}(\mu\mid\HB)]
+
\E[c_{\mathscr X}],
\label{eq:cross-moment-decomposition}
\end{align}
and therefore
\begin{equation}
\boxed{
V_B(\HB)-\Vpol(\pi_f)
\le
\sqrt{M_\times(f)}.
}
\label{eq:bridge}
\end{equation}
Consequently, for any policy $\pi_G$ measurable with respect to a refinement $\GB\supseteq\HB$,
\begin{equation}
\Vpol(\pi_G)-V_B(\HB)
\ge
\bigl\{\Vpol(\pi_G)-\Vpol(\pi_f)\bigr\}
-
\sqrt{M_\times(f)}.
\label{eq:bridge-policy-improvement}
\end{equation}
\end{theorem}

The decomposition shows why the bridge is conservative.  Even when $f=m_{\HB}$, persistent hidden heterogeneity contributes the nonnegative floor
\begin{equation}
V_{\mathrm{hid}}
:=
\E[\operatorname{Var}(\mu\mid\HB)].
\end{equation}
The bridge is therefore sufficient rather than necessary: not clearing it does not imply that the active value is below the coarse-information Bayes value.

\begin{corollary}[Conditional bridge]
Let $A\in\HB$ with $\Pp(A)>0$.  Replacing all expectations by expectations conditional on $A$ gives
\begin{equation}
V_B(\HB;A)-\Vpol(\pi_f;A)
\le
\sqrt{M_\times(f;A)},
\end{equation}
where
\begin{equation}
M_\times(f;A)
=
\E[(D_A-f)(D_B-f)\mid A].
\label{eq:conditional-cross-moment}
\end{equation}
\end{corollary}

\section{Adaptive-depth certification details}
\label{app:adaptive-cert}

This appendix records the structural certification layer used to interpret adaptive revelation depth. The generic risk-calibration tools are established prior art; the purpose here is to state how they interact with productive-prefix reuse and terminal decision severity in the present execution technology.

Let $S\in\{0,1\}$ denote an H4 early-stop decision, let $F=\ind\{\pi_4\neq\pi_8\}$, and define $Y=Q_{\pi_8}-Q_{\pi_4}$. Each H4 stop saves compute value $c=4\lambda_0$ relative to always-H8.

\begin{proposition}[Productive-prefix decision invariance]
\label{prop:productive-invariance}
Under the declared same-path promotion technology, if $F=0$ then $Y=0$. For the binary action menu,
\begin{equation}
Y=D\left(\ind\{\pi_8=\XI\}-\ind\{\pi_4=\XI\}\right),
\qquad |Y|=|D|F.
\label{eq:productive-invariance}
\end{equation}
\end{proposition}

\begin{proposition}[Risk--severity factorization]
\label{prop:risk-severity}
Let $\rho=\Pr(S=1)$ and $r=\Pr(S=1,F=1)$. When $r>0$, define the signed stopped-flip severity $\eta_s=\E[Y\mid S=1,F=1]$ and the positive-harm severity $\eta_+=\E[Y_+\mid S=1,F=1]$, where $Y_+=\max(Y,0)$. When $r=0$, set $\eta_s=\eta_+=0$ by convention. Then
\begin{equation}
\boxed{\Delta V(S)=\E[S(c-Y)]=c\rho-r\eta_s.}
\label{eq:risk-severity-factorization}
\end{equation}
Moreover,
\begin{equation}
\Delta V(S)\ge c\rho-r\eta_+.
\label{eq:risk-severity-positive}
\end{equation}
\end{proposition}

\begin{theorem}[Risk-only impossibility and the harm-tail boundary]
\label{thm:risk-only-impossibility}
Let $E=\{S=1,F=1\}$ and define the stopped positive-harm mass
\begin{equation}
H_+:=\E[Y_+\ind_E]\in[0,\infty].
\label{eq:stopped-harm-mass}
\end{equation}
Then
\begin{equation}
H_+
=
\int_0^\infty \Pr(E,\,Y_+>t)\,dt,
\qquad
\Delta V(S)\ge c\rho-H_+.
\label{eq:harm-tail-integral}
\end{equation}
Moreover, fix any $0<r\le\rho\le1$. For every $M>0$ there exists a joint law of $(S,F,Y)$ satisfying productive-prefix decision invariance, with $\Pr(S=1)=\rho$, $\Pr(E)=r$, and finite $\E[Y_+]$, such that
\begin{equation}
\Delta V(S)<-M.
\label{eq:risk-only-unbounded-below}
\end{equation}
Hence, whenever nonzero stop--flip risk is permitted, exact knowledge of $(\rho,r)$ alone gives no finite distribution-robust lower bound on expected net utility over a class with unrestricted stopped-flip severity. Any positive expected-utility certificate must additionally control $H_+$, or a sufficient severity/tail surrogate for it.
\end{theorem}

\begin{corollary}[Minimal sufficient harm-tail bridges]
\label{cor:severity-bridges}
The positive-harm component left uncontrolled by stop--flip risk is summarized by the integrated stopped-harm tail $H_+$ in Equation~\eqref{eq:stopped-harm-mass}. The following declared conditions are sufficient ways to control it for a lower utility certificate.

If $Y_+\le B_+$ almost surely, then
\begin{equation}
\Delta V(S)\ge c\rho-B_+r.
\label{eq:uniform-severity-bridge}
\end{equation}
If instead $q>1$ and $\E[|D|^q]\le M_q$, then H\"older's inequality gives
\begin{equation}
\Delta V(S)\ge c\rho-M_q^{1/q}r^{1-1/q}.
\label{eq:moment-severity-bridge}
\end{equation}
More generally, if an integrable envelope $u:[0,\infty)\to[0,\infty)$ satisfies
\begin{equation}
\Pr(E,\,Y_+>t)\le u(t)
\qquad\text{for all }t\ge0,
\end{equation}
then
\begin{equation}
\Delta V(S)
\ge
c\rho-\int_0^\infty u(t)\,dt.
\label{eq:tail-envelope-bridge}
\end{equation}
Thus a finite-sample event $\rho\ge L_\rho$, $r\le U_r$ yields certified lower bounds $cL_\rho-B_+U_r$ or $cL_\rho-M_q^{1/q}U_r^{1-1/q}$ under the corresponding declared severity class; an independently valid upper bound $U_H$ on $H_+$ yields the assumption-matched bound $cL_\rho-U_H$.
\end{corollary}

Theorem~\ref{thm:risk-only-impossibility} is a control-specific boundary, not a claim that bounded-risk calibration is impossible. The Bernoulli event $E$ can be calibrated without utility-tail assumptions; what cannot be obtained from that event probability alone is a nontrivial lower bound on an unbounded mean severity. At the statistical level, this distinction is aligned with classical nonparametric impossibility results for unrestricted mean inference \citep{bahadur1956nonexistence}.

\paragraph{Bounded decision-instability calibration.}
The loss $L=\ind\{S=1,F=1\}$ is Bernoulli. Hence fixed-sequence calibration or other established bounded-risk procedures such as Learn-then-Test and Conformal Risk Control can be used prospectively once the risk model, threshold family, calibration data, and risk target are fixed \citep{angelopoulos2025ltt,angelopoulos2024crc}. This controls how often early stopping can change the deeper action, not how severe those changed decisions are.

\paragraph{Fixed-data risk--severity diagnostic.}
A fixed-data nested evaluation illustrates the distinction. A two-coordinate H4 flip-risk score is trained on one collection block, calibrated on one future bank of the other block to target 2\% joint stop-and-flip risk, and checked on the remaining future bank. Table~\ref{tab:risk-severity-diagnostic} reports the held-out outcomes. The first direction has positive realized net value; the second has a lower flip rate but negative realized net value because its single harmful flip has substantially larger severity. This is a diagnostic validation of Equation~\eqref{eq:risk-severity-factorization}, not a prospective controller claim.

\begin{table}[H]
\centering
\caption{Held-out risk--severity diagnostic for a derived risk-calibrated stopping rule. Net value is relative to always-H8.}
\label{tab:risk-severity-diagnostic}
\small
\begin{tabular}{lrrrrr}
\toprule
Direction & Stop rate & Stop--flip rate & Harm events & Max harm severity & Net value\\
\midrule
A $\to$ B & 26.67\% & 0.833\% & 1 & $2.083\times10^{-4}$ & $+1.186\times10^{-5}$\\
B $\to$ A & 11.67\% & 0.417\% & 1 & $2.010\times10^{-3}$ & $-4.331\times10^{-6}$\\
\bottomrule
\end{tabular}
\end{table}

The diagnostic shows that a probability budget is not a utility budget: decision-instability risk controls how often early stopping can differ from H8, while the integrated harm tail controls how costly those differences are. Theorem~\ref{thm:risk-only-impossibility} makes this separation sharp. A fully prospective expected-utility certificate therefore needs both a bounded-risk calibration and a predeclared severity, moment, or integrable tail class.

\section{Proofs}

\subsection{Proof of Proposition~\ref{prop:factorization-criterion}}
Let $M_a=\E[Q_a\mid\GB]$.  Because $\mathcal A$ is finite, a $\GB$-Bayes action exists after fixing a measurable tie rule.  If a $\GB$-Bayes action $\pi_H$ is $\HB$-measurable, then
\begin{equation}
V_B(\GB)=\E[M_{\pi_H}]=\E[Q_{\pi_H}]\le V_B(\HB)\le V_B(\GB),
\end{equation}
so equality holds.  Conversely, let $\pi_H^\star$ be an $\HB$-Bayes policy.  If $V_B(\HB)=V_B(\GB)$, then
\begin{equation}
0=V_B(\GB)-\Vpol(\pi_H^\star)=\E\!\left[\max_a M_a-M_{\pi_H^\star}\right].
\end{equation}
The integrand is nonnegative, hence it vanishes almost surely and $\pi_H^\star$ is also $\GB$-Bayes optimal.

\subsection{Proof of Proposition~\ref{prop:df}}
By definition,
\begin{align}
\mathcal O_t^{\rm DF}
&=L_t+\max_a\{\nu_t(a)+\gamma\omega_t(a)\}-\nu_t^\star\\
&=L_t+\max_a\{\gamma\omega_t(a)-(\nu_t^\star-\nu_t(a))\}\\
&=L_t+\max_a\{\gamma\omega_t(a)-d_t(a)\}.
\end{align}
The maximum term is nonnegative because any coarse-optimal action has $d_t(a)=0$ and $\omega_t(a)\ge0$.  Since $L_t\ge0$, the sum is zero if and only if both $L_t=0$ and every term inside the maximum is nonpositive.

\subsection{Proof of Proposition~\ref{prop:productive-budget}}
Temporary probing spends $mc$ updates across $m$ candidates and then restarts the selected action for the full horizon $H$, so $K_F(c)=mc+H$.  Probe-and-promote revelation spends $mh$ updates across candidates and then only $H-h$ further updates on the selected path, so $K_{\AR}(h)=H+(m-1)h$.  Equating the budgets gives $(m-1)h=mc$.

\subsection{Proof of Proposition~\ref{prop:learnability-ledger}}
Let $V_{\Pi_s}^{\star}=V_{P_s}^{\star}-R_s^{\rm app}$ be the best value in the learner class and let $W_{s,n}=V_{\Pi_s}^{\star}-R_{s,n}^{\rm est}$ be the learned gross value.  Since $V_{P_s}^{\star}=V_B(\HB)+\Delta_s^{\rm rev}+\Delta_s^{\rm prom}$ and $W_{s,n}^{\rm net}=W_{s,n}-C_s$, substitution yields Equation~\eqref{eq:learnability-ledger}.

\subsection{Proof of Corollary~\ref{cor:frontier-ledger}}
Using $V_P(\hat\pi_A)=V_B(\GB)-R_A+P_A$ and $V_R(\hat\pi_F)=V_B(\HB)-R_F$,
\begin{equation}
V_P(\hat\pi_A)-V_R(\hat\pi_F)
=[V_B(\GB)-V_B(\HB)]+P_A+R_F-R_A.
\end{equation}
Subtracting priced resource costs gives the net-value version.

\subsection{Proof of Theorem~\ref{thm:alias-cell}}
Relative to $\KEEP$, the refined Bayes increment is $\E[(m_G)_+]$ and the coarse Bayes increment is $\E[(m_H)_+]$.  Conditioning the refined increment on $\HB$ gives $a_H$.  Since
\begin{equation}
m_H
=\E[m_G\mid\HB]
=a_H-b_H,
\end{equation}
we have pointwise
\begin{equation}
a_H-(m_H)_+
=a_H-(a_H-b_H)_+
=\min\{a_H,b_H\}.
\end{equation}
Taking expectations proves Equation~\eqref{eq:alias-cell-gap}.  On the event $A$ in Equation~\eqref{eq:alias-sign-mass},
\begin{equation}
a_H\ge\delta\alpha,
\qquad
b_H\ge\delta\beta,
\end{equation}
so $\min\{a_H,b_H\}\ge\delta\min(\alpha,\beta)$ there.  Integrating over $A$ gives Equation~\eqref{eq:alias-lower-bound}.

\subsection{Proof of Proposition~\ref{prop:local-alias}}
By the constant-rank theorem, the local level set $\Phi_c^{-1}(\Phi_c(x_0))$ is a submanifold with tangent space $\ker D\Phi_c(x_0)$.  Hence there is a $C^1$ curve $\gamma(s)$ in that level set with $\gamma(0)=x_0$ and $\gamma'(0)=v$.  Taylor expansion gives
\begin{equation}
g(\gamma(s))=s\,Dg(x_0)v+o(s),
\end{equation}
so for sufficiently small positive and negative $s$ the terminal gap has opposite signs.  Likewise
\begin{equation}
Y_h(\gamma(s))=Y_h(x_0)+s\,DY_h(x_0)v+o(s),
\end{equation}
which changes to first order because $DY_h(x_0)v\neq0$.

\subsection{Proof of the replicate-denoised bridge}
By iterated expectation,
\begin{align}
M_\times(f)
&=\E\!\left[\E[(D_A-f)(D_B-f)\mid\mathscr X]\right]\\
&=\E[(\mu-f)^2]+\E[c_{\mathscr X}].
\end{align}
Since $f$ is $\HB$-measurable and $m_{\HB}=\E[\mu\mid\HB]$, the conditional Pythagorean identity gives
\begin{equation}
\E[(\mu-f)^2]
=
\E[(m_{\HB}-f)^2]
+\E[\operatorname{Var}(\mu\mid\HB)].
\end{equation}
For binary actions with the declared tie rule $\pi_f=\ind\{f>0\}$, the regret relative to the $\HB$-Bayes rule $\pi_{m}=\ind\{m_{\HB}>0\}$ is
\begin{equation}
V_B(\HB)-\Vpol(\pi_f)
=
\E\!\left[|m_{\HB}|\,\ind\{\pi_f\neq\pi_m\}\right].
\end{equation}
On action disagreement, $|m_{\HB}|\le|m_{\HB}-f|$; ties at $m_{\HB}=0$ contribute zero regret.  Cauchy--Schwarz therefore yields
\begin{equation}
V_B(\HB)-\Vpol(\pi_f)
\le\E|m_{\HB}-f|
\le\sqrt{\E[(m_{\HB}-f)^2]}
\le\sqrt{M_\times(f)}.
\end{equation}
Adding and subtracting $\Vpol(\pi_f)$ proves the policy-improvement inequality.  Conditioning throughout on $A\in\HB$ proves the conditional version.

\subsection{Proof of Proposition~\ref{prop:static-dynamic-embedding}}
By the tower property,
\begin{align}
\E[\mathcal G_{h\to h'}]
&=\E\!\left[\E[C_{h'}\mid\mathscr F_h]-C_h\right]\\
&=\E[C_{h'}]-\E[C_h]\\
&=V_B(\mathscr F_{h'})-V_B(\mathscr F_h).
\end{align}
The final equality is the definition of Bayes value under each information state.

\subsection{Proof of Proposition~\ref{prop:adaptive-bellman}}
At the final admissible depth $h_J$, no deeper revelation is available, so $V_J^{\rm dyn}=C_{h_J}$. At an earlier depth, the controller has exactly two admissible choices in the stated finite-depth problem: commit immediately for value $C_{h_j}$, or buy the next refinement, pay $c_j$, and receive conditional expected value $\E[V_{j+1}^{\rm dyn}\mid\mathscr F_{h_j}]$. Taking the larger value gives Equation~\eqref{eq:adaptive-bellman}. Backward induction completes the recursion.

\subsection{Proof of Proposition~\ref{prop:vor-stopping}}
Given $\mathscr F_h$, the conditional net increment from continuing is $G_{h\to h'}-c_{h\to h'}$; hence continuation is optimal exactly when this quantity is positive.

\subsection{Proof of Theorem~\ref{thm:scalar-control-factorization}}
Write $\Gamma=\Gamma_{h\to h'}$ and condition on the scalar summary $S_h$.  Since
\begin{equation}
\Gamma=\Gamma_+-(-\Gamma)_+,
\end{equation}
we have
\begin{equation}
\E[\Gamma\mid S_h]=a(S_h)-b(S_h).
\end{equation}
Therefore
\begin{align}
V_{\rm meta}(\mathscr F_h)-V_{\rm meta}(S_h)
&=\E[a(S_h)]-\E[(a(S_h)-b(S_h))_+]\\
&=\E\!\left[a(S_h)-(a(S_h)-b(S_h))_+\right]\\
&=\E[\min\{a(S_h),b(S_h)\}],
\end{align}
which proves Equation~\eqref{eq:scalar-control-gap}.  Because $a,b\ge0$, the gap is zero if and only if $\min\{a,b\}=0$ almost surely.  For an integrable random variable, $a(S_h)=0$ exactly when $\Pr(\Gamma>0\mid S_h)=0$ almost surely on that scalar fiber, and similarly $b(S_h)=0$ exactly when $\Pr(\Gamma<0\mid S_h)=0$.  Thus equality holds if and only if no positive-mass scalar fiber contains both strictly positive and strictly negative net continuation gains.  Equivalently, there exists a full-information optimal meta-action whose tie assignment at $\Gamma=0$ is measurable with respect to $S_h$; that optimal meta-action then factorizes through $S_h$. This is precisely value-preserving decision factorization for the two-action meta-menu $\{\mathrm{STOP},\mathrm{CONTINUE}\}$.

\subsection{Proof of Theorem~\ref{thm:two-coordinate-revelation}}
By symmetry and integrability of $Z$, $\E[Z]=0$. If $m_h\ge0$, then
\begin{align}
\E[(m_h+\sigma_h Z)_+\mid\mathscr F_h]-m_h
&=\E[(-m_h-\sigma_h Z)_+\mid\mathscr F_h]\\
&=\sigma_h\E[(Z-m_h/\sigma_h)_+],
\end{align}
where the last equality uses symmetry. If $m_h<0$, the coarse positive-part value is zero and
\begin{equation}
\E[(m_h+\sigma_h Z)_+\mid\mathscr F_h]
=\sigma_h\E[(Z-|m_h|/\sigma_h)_+].
\end{equation}
This proves Equation~\eqref{eq:two-coordinate-value}. Since $\psi'(t)=-\Pr(Z>t)$ at continuity points,
\begin{equation}
\frac{\partial\mathcal R}{\partial |m_h|}=\psi'(t)=-\Pr(Z>t).
\end{equation}
For $t=|m_h|/\sigma_h$,
\begin{equation}
\frac{\partial\mathcal R}{\partial\sigma_h}
=\psi(t)-t\psi'(t)=\psi(t)+t\Pr(Z>t)\ge0
\end{equation}
with strict inequality whenever the refinement retains positive tail mass beyond $t$.

\subsection{Proof of Corollary~\ref{cor:conditional-revealability}}
If $\sigma_h=g(M_h)$ on $E$, substitution into Equation~\eqref{eq:two-coordinate-value} gives
\begin{equation}
\mathcal R_{h\to h'}=g(M_h)\psi\!\left(M_h/g(M_h)\right)=\widetilde r(M_h)
\end{equation}
on $E$, proving scalar sufficiency for the one-step continuation value. A threshold representation is a stronger statement and requires the resulting scalar value function to be monotone relative to cost.

For the second part, condition on $M_h=m$. By assumption, the conditional distribution of $\sigma_h$ is nondegenerate on a set of $m$ values with positive probability, and $s\mapsto r(m,s)$ is strictly increasing over that conditional support. A strictly monotone transform of a nondegenerate random variable is nondegenerate. Equation~\eqref{eq:two-coordinate-value} identifies this transform with $\mathcal R_{h\to h'}$, so the conditional law of $\mathcal R_{h\to h'}$ given $M_h$ is nondegenerate on that set and no $M_h$-measurable scalar function can equal the continuation value almost surely there.

\subsection{Proof of Corollary~\ref{cor:cost-adjusted-revealability-crossing}}
For fixed $M_h=m$, let
\begin{equation}
\Gamma=\mathcal R_{h\to h'}-c=r(m,\sigma_h)-c.
\end{equation}
Theorem~\ref{thm:scalar-control-factorization} gives scalar sufficiency exactly when, conditional on almost every $m$, $\Gamma$ does not have both strictly positive and strictly negative mass. This is equivalent to Equation~\eqref{eq:crossing-sufficiency}: either there is no strictly positive mass, or there is no strictly negative mass. Zero-gain ties may be assigned to either optimal side without changing value. Under continuity and strict monotonicity, $r(m,s)\le c$ for $s\le\sigma_c(m)$ and $r(m,s)\ge c$ for $s\ge\sigma_c(m)$ on the relevant support (with equality possible only at the crossing), which gives the stated support-side sufficient condition.

For the second statement, Equation~\eqref{eq:revealability-crossing} implies that on a positive-probability set of margin fibers there is positive conditional probability of both $\Gamma<0$ and $\Gamma>0$. Hence both conditional positive-part expectations $a(M_h)$ and $b(M_h)$ from Theorem~\ref{thm:scalar-control-factorization} are strictly positive on that set. Their minimum therefore has strictly positive expectation, so Equation~\eqref{eq:scalar-control-gap} yields
\begin{equation}
V_{\rm meta}(\mathscr F_h)-V_{\rm meta}(M_h)>0.
\end{equation}

\subsection{Proof of Proposition~\ref{prop:revealability-scale-identification}}
Equation~\eqref{eq:location-scale-refinement} gives $m_{h'}-m_h=\sigma_h Z$. Because $\sigma_h$ is $\mathscr F_h$-measurable and $Z$ is independent of $\mathscr F_h$,
\begin{equation}
\E[|m_{h'}-m_h|\mid\mathscr F_h]
=\sigma_h\E|Z|.
\end{equation}
If $\E[Z^2]=1$, the same argument yields
\begin{equation}
\E[(m_{h'}-m_h)^2\mid\mathscr F_h]
=\sigma_h^2\E[Z^2]=\sigma_h^2.
\end{equation}

\subsection{Proof of Proposition~\ref{prop:productive-invariance}}
If $F=0$, then $\pi_4=\pi_8$ and both policies promote the same already-tested path under the declared same-path technology. Their H12 terminal outcome is therefore identical, so $Y=Q_{\pi_8}-Q_{\pi_4}=0$. For the binary action menu,
\begin{equation}
Q_{\pi_j}=Q_{\KEEP}+D\ind\{\pi_j=\XI\},
\end{equation}
for $j\in\{4,8\}$. Subtracting the two identities gives Equation~\eqref{eq:productive-invariance}; because the two action indicators differ exactly when $F=1$, $|Y|=|D|F$.

\subsection{Proof of Proposition~\ref{prop:risk-severity}}
By Proposition~\ref{prop:productive-invariance}, $SY=0$ outside $E=\{S=1,F=1\}$. Hence
\begin{align}
\Delta V(S)
&=\E[S(c-Y)]\\
&=c\Pr(S=1)-\E[Y\ind_E]\\
&=c\rho-r\E[Y\mid E]
=c\rho-r\eta_s,
\end{align}
when $r>0$. Since $Y\le Y_+$ pointwise,
\begin{equation}
\E[Y\ind_E]\le\E[Y_+\ind_E]=r\eta_+,
\end{equation}
which yields Equation~\eqref{eq:risk-severity-positive}. The case $r=0$ reduces to $\Delta V(S)=c\rho$.

\subsection{Proof of Theorem~\ref{thm:risk-only-impossibility}}
For the nonnegative random variable $Y_+\ind_E$, the layer-cake identity gives
\begin{equation}
\E[Y_+\ind_E]
=\int_0^\infty \Pr(Y_+\ind_E>t)\,dt
=\int_0^\infty \Pr(E,\,Y_+>t)\,dt.
\end{equation}
Combining this identity with Equation~\eqref{eq:risk-severity-positive} proves Equation~\eqref{eq:harm-tail-integral}.

For the impossibility statement, fix $0<r\le\rho\le1$ and $M>0$. Choose a joint law with
\begin{equation}
\Pr(S=1,F=1)=r,
\qquad
\Pr(S=1,F=0)=\rho-r,
\end{equation}
and allocate the remaining probability to $S=0,F=0$. Set $Y=0$ whenever $F=0$ and set
\begin{equation}
Y=K:=\frac{M+c\rho+1}{r}
\end{equation}
on $E=\{S=1,F=1\}$. This law satisfies productive-prefix decision invariance and has finite $\E[Y_+]=rK$. Its net value is
\begin{equation}
\Delta V(S)=c\rho-rK=-M-1<-M.
\end{equation}
Because $M$ is arbitrary while $(\rho,r)$ are fixed, no finite lower bound depending only on $(c,\rho,r)$ can hold uniformly over unrestricted stopped-flip severities.

\subsection{Proof of Corollary~\ref{cor:severity-bridges}}
If $Y_+\le B_+$ almost surely, then
\begin{equation}
H_+=\E[Y_+\ind_E]\le B_+\Pr(E)=B_+r,
\end{equation}
which gives Equation~\eqref{eq:uniform-severity-bridge}. If $\E[|D|^q]\le M_q$ with $q>1$, Proposition~\ref{prop:productive-invariance} and H\"older's inequality yield
\begin{align}
H_+
&\le\E[|D|\ind_E]\\
&\le \E[|D|^q]^{1/q}\Pr(E)^{1-1/q}\\
&\le M_q^{1/q}r^{1-1/q},
\end{align}
proving Equation~\eqref{eq:moment-severity-bridge}. Finally, Equation~\eqref{eq:harm-tail-integral} and the assumed tail envelope give
\begin{equation}
H_+
\le\int_0^\infty u(t)\,dt,
\end{equation}
which proves Equation~\eqref{eq:tail-envelope-bridge}. Substituting simultaneous confidence bounds for $\rho$, $r$, or $H_+$ gives the stated finite-sample lower certificates.

\section{Notation and decision objects}
\label{app:notation}
\begin{center}
\begin{tabularx}{\linewidth}{>{\raggedright\arraybackslash}p{0.20\linewidth}X}
\toprule
Symbol & Meaning\\
\midrule
$\HB,\GB$ & Coarse legal information and a refinement\\
$V_B(\mathcal I)$ & Bayes value under information $\mathcal I$\\
$\mathscr F_h, C_h$ & Revelation filtration and Bayes commit value at depth $h$\\
$Q_a$ & Terminal utility for action $a$\\
$D$ & Binary terminal gap $Q_{\XI}-Q_{\KEEP}$\\
$\Delta^{\rm rev}$ & Pure information-refinement value\\
$\Delta^{\rm prom},P_A$ & Productive path-reuse / promotion value\\
$R^{\rm app},R^{\rm est}$ & Approximation and finite-sample estimation regret\\
$C_s$ & Acquisition cost\\
$\Phi_c$ & Complete legal short-probe frontier information map\\
$Y_h$ & Deeper future-learning observation\\
$\tau^\star(v)$ & First depth at which direction $v$ is revealed\\
$M_\times(f)$ & Cross-replicate residual moment used in the optional conservative bridge\\
$A$ & Fixed H4 frontier-ambiguity event in the 432-anchor structural analysis\\
$\Delta_{\rm restart},\Delta_{\rm path}$ & Common-restart policy difference and productive path-reuse value\\
$\mathcal G_{h\to h'}$ & Oracle conditional Bayes value of purchasing a deeper information refinement\\
$G_{h\to h'}$ & Policy gain from deeper revelation, conditional on $\mathscr F_h$\\
$\Gamma_{h\to h'}$ & Cost-adjusted conditional continuation gain $G_{h\to h'}-c_{h\to h'}$\\
$\sigma_h$ & State-dependent revealability scale in the location--scale stopping model\\
$\sigma_c(m)$ & Critical revealability at which the priced \textsc{Stop}/\textsc{Continue} decision changes for margin $m$\\
$R_h^{(\mathcal M)}$ & Legal model-specific empirical proxy for revealability, fixed without target outcomes\\
$S,F$ & Early-stop indicator and shallow/deep action-instability indicator\\
$\rho,r$ & Stop probability and joint stop--flip probability in the risk--severity bridge\\
\bottomrule
\end{tabularx}
\end{center}

\section{Statistical estimands and simultaneous inference}
\label{app:inference}

This appendix collects the inferential rules used by the empirical claims. The unit of resampling and inference is the \emph{exact anchor}, not a bank-level realization, readout, or policy action. When two conditionally independent future banks are available for one anchor, bank-specific contributions are averaged within the anchor before estimating a population mean.

\paragraph{One-sided mean lower bounds.}
For exact-anchor contributions $Z_1,\ldots,Z_n$, let $\bar Z$ and $s_Z$ denote the sample mean and standard deviation. The reported one-sided Student-$t$ lower bound is
\begin{equation}
\mathrm{LCB}_{t,1-\alpha}
=
\bar Z-t_{1-\alpha,n-1}\frac{s_Z}{\sqrt n}.
\end{equation}
The percentile bootstrap resamples exact anchors with replacement. Unless otherwise stated, 50,000 draws are used and the lower bound is the empirical $\alpha$ quantile of the bootstrap means. These uncertainty summaries target the declared exact-anchor sampling design and panel-generating regime; they are not universal guarantees over arbitrary model families, task streams, or deployment distributions.

\paragraph{Finite-library multiplicity.}
The two 336-anchor Qwen finite-library panels each test the same 22 prespecified current-information comparator components. Every component must have a positive point margin, one-sided $t$ lower bound, and exact-anchor bootstrap lower bound. In addition, a 200,000-draw max-$t$ procedure supplies simultaneous 95\% lower bounds across all 22 components. The two panels are independent replications and are never pooled for this conjunction.

\paragraph{Multiplicity scope.}
Multiplicity is controlled within each prespecified family to which a conjunction or model-selection claim is attached: the 22-component current-information library and, separately, the two-architecture Mistral adaptive family. The manuscript does not assert a single paper-wide familywise-error guarantee across heterogeneous theory-targeted estimands, which answer distinct scientific questions and are reported with their own declared inferential units and lower bounds.

\paragraph{Equal-compute decomposition.}
The 432-anchor common-restart comparison and each 336-anchor productive-path panel are independent. Equation~\eqref{eq:end-to-end-decomp} is therefore evaluated once with productive-path panel I and once with panel II. Standard errors use the independence-aware Welch--Satterthwaite construction; the two productive-path panels are not pooled.

\paragraph{Mistral two-bank evaluation.}
The fixed-depth and equal-compute Mistral estimands average banks A and B within each of 480 exact anchors before $t$ or bootstrap inference. The adaptive target analysis likewise forms $Z_i^{AB}=(Z_{i,A}+Z_{i,B})/2$ before inference.

\paragraph{Finite architecture family for Mistral adaptive control.}
Scalar and two-coordinate continuation architectures were both specified in the Qwen adaptive-control analysis before the Mistral target analysis. Their Mistral parameters were fit using only the separate 336-anchor development panel. To protect the target-panel claim against selecting between these two architectures, they are treated as a two-element finite family. The scalar target-panel point estimate is
\begin{equation}
\widehat{\Delta V}_{\rm scalar}^{\mathrm M}=1.13877630257\times10^{-5}.
\end{equation}
Its ordinary one-sided 95\% $t$ and bootstrap lower bounds are $2.37141363069\times10^{-6}$ and $5.41900958243\times10^{-6}$. Bonferroni familywise 95\% bounds use one-sided level $1-0.05/2=0.975$ for each architecture and remain positive for the scalar rule:
\begin{equation}
\mathrm{LCB}^{\rm FWER}_{t}=6.37735915928\times10^{-7},
\qquad
\mathrm{LCB}^{\rm FWER}_{\rm boot}=5.23837592968\times10^{-6}.
\end{equation}
Later exploratory revealability proxies are not used to support this target-panel result.

\paragraph{Bounded stop--flip risk.}
For a fixed stop rule, the joint event $\{S=1,F=1\}$ is Bernoulli. Reported risk certificates are one-sided Bernoulli--KL upper confidence bounds computed bankwise under the prespecified calibration role. These bounds concern the probability of changing the deeper action; Theorem~\ref{thm:risk-only-impossibility} explains why they do not by themselves bound the utility severity of those changes.

\section{Experimental protocol details}

\paragraph{Qwen comparative panel.}
The Qwen comparative panel contains 432 exact anchors disjoint from development and from the two finite-library panels, with three prespecified task-stream strata of 144 anchors each and nine-position balance. The active policy, strengthened DynaMiCS-style frontier comparator, 20-update policy-visible compute contract, and ambiguity threshold
\begin{equation}
\tau_F=0.00052099142651661841
\end{equation}
were fixed before terminal evaluation. Each anchor contains both $\KEEP$ and $\XI$ H12 outcomes in two conditionally independent terminal banks, $T_1$ and $T_2$.

\paragraph{Mistral development and target panels.}
Mistral-specific numerical heads are fit on a separate 336-anchor development panel. The independent 480-anchor target panel uses model revision \texttt{c03fc1dabc3d31b96271626f15a76a6779fb4037} of \texttt{mistralai/Mistral-7B-v0.3}, two independent future banks, the same $\KEEP/\XI$ action menu, H4/H8/H12 geometry, and 20-update policy-visible compute contract. The H4 ambiguity threshold $0.00236920914414$ is estimated from the development panel and fixed for target evaluation.

\phantomsection
\section*{References}
\addcontentsline{toc}{section}{References}
\renewcommand{\bibsection}{}
\setlength{\bibsep}{0.55pt plus 0.2pt minus 0.1pt}
\fontsize{8.35}{9.7}\selectfont
\raggedright
\bibliographystyle{plainnat}
\bibliography{references}
\end{document}